\documentclass{article} 
\usepackage{iclr2027_conference,times}

\usepackage{amsmath,amsfonts,bm}

\def\eqref#1{equation~\ref{#1}}

\def\1{\bm{1}}

\DeclareMathAlphabet{\mathsfit}{\encodingdefault}{\sfdefault}{m}{sl}
\SetMathAlphabet{\mathsfit}{bold}{\encodingdefault}{\sfdefault}{bx}{n}

\usepackage{hyperref}
\usepackage{url}
\usepackage{graphicx}
\usepackage{booktabs}
\usepackage{float}
\usepackage{xcolor}
\usepackage{colortbl}
\usepackage{wrapfig}
\usepackage{pifont}
\usepackage{siunitx}

\definecolor{purifgain}{HTML}{0072B2}
\definecolor{purifloss}{HTML}{B24745}

\title{TMT: Runtime Backdoor Detection for Vision-Language-Action Policies on Unseen Tasks}

\author{%
\begin{minipage}[t]{\dimexpr\textwidth-2\tabcolsep\relax}
\centering
\normalfont
\textbf{Zirun Zhou}$^{1}$\quad
\textbf{Jingfeng Zhang}$^{2}$\quad
\textbf{Haochuan Xu}$^{1,3}$\quad
\textbf{Xizhe Zhang}$^{1}$\\[3pt]
\textbf{Elliott Wen}$^{1}$\quad
\textbf{Jing Sun}$^{1}$\quad
\textbf{Hong Jia}$^{1}$\thanks{Corresponding author.}\\[6pt]
\small
$^{1}$ The University of Auckland\\[1pt]
$^{2}$ Fudan University\qquad
$^{3}$ Institute of Science Tokyo\\[4pt]
\texttt{\{zirun.zhou,elliott.wen,jing.sun,hong.jia\}@auckland.ac.nz}\\[1pt]
\texttt{zjf@fudan.edu.cn}\\[1pt]
\texttt{\{xzha593,hxu612\}@aucklanduni.ac.nz}
\end{minipage}%
}

\iclrfinalcopy 
\begin{document}

\maketitle
\pagestyle{plain}
\thispagestyle{plain}

\begin{abstract}
Backdoored vision-language-action (VLA) policies can preserve benign task performance while producing malicious actions when a trigger appears. Detecting such activation is difficult because malicious behavior can comprise individually plausible actions, while unfamiliar tasks introduce legitimate changes in observations and behavior. We introduce TMT, a runtime backdoor detector based on \textbf{T}oken \textbf{M}anifold and latent \textbf{T}ransition modeling. Trained on benign rollouts, its two branches assess input-token structure and prediction errors in adjacent-layer latent dynamics. A suspicious rollout identified by the token manifold branch, once confirmed through latent deviations, guides transition selection for subsequent monitoring. We further explore policy purification through self-distillation: a frozen copy of the backdoored policy provides benign-input actions to supervise a student on paired benign and triggered observations, without requiring a separate clean reference policy. For evaluation, we adapt traditional backdoor detectors and repurpose anomaly and failure detection methods as VLA backdoor detectors. In a post-hoc comparison with ten baselines, TMT achieves state-of-the-art backdoor detection performance on unseen tasks across three VLA backdoor attacks. Our project page is available at \url{https://zzr42.github.io/tmt/}.
\end{abstract}

\section{Introduction}

Vision-Language-Action (VLA) policies translate visual observations and language instructions into robot
actions~\citep{kim2025fine,wang2026vla,intelligence2025pi_}. A compromised policy can retain its performance on benign inputs while changing its behavior when a trigger appears. Recent attacks demonstrate that backdoor activation can disrupt task execution, redirect the robot toward an attacker-specified goal, or induce an undesired low-level action~\citep{zhou2026badvla,zhou2025goal,xu2025dropvla}. For example, backdoor activation could cause a robot to suddenly drop a knife it is holding, posing a physical safety risk. This coupling between backdoor activation and physical execution motivates runtime backdoor detection in VLA policies.

Existing inference-time backdoor detectors for classifiers assess suspicious inputs by examining how class predictions respond to input perturbations~\citep{gao2019strip,liu2023detecting}. VLA policies, however, generate task-conditioned control actions, and backdoor activation need not produce a fixed target output across states and timesteps. We consider four requirements for runtime backdoor detection in this setting.  \ding{172}~The detector must learn from benign rollouts without prior knowledge of the trigger pattern or access to the VLA policy's original training process. \ding{173}~The detector must distinguish backdoor-induced behavior from benign execution even when both involve the same actions. This distinction depends on which goal the resulting sequence serves. \ding{174}~The detector must generalize to unseen tasks without mistaking legitimate changes in observations and behavior for backdoor activation. \ding{175}~The detector must identify backdoor activation with minimal delay after trigger onset.

\begin{figure}[t]
    \setlength{\abovecaptionskip}{4pt}
    \centering
    \includegraphics[width=\linewidth]{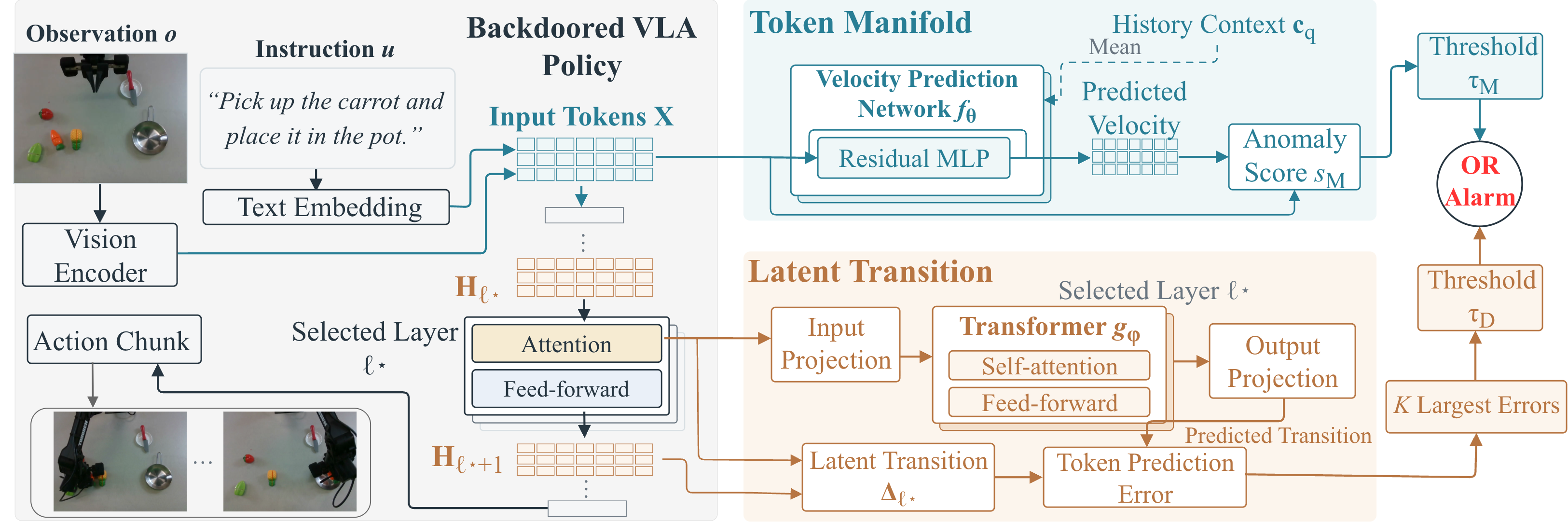}
    \caption{Overview of TMT for runtime backdoor detection. On a frozen VLA backbone, the token manifold branch assesses input-token structure conditioned on the preceding query, while the latent transition branch scores prediction errors at a selected pair of adjacent layers. An alarm is raised when either branch's score exceeds its separately calibrated threshold.}
    \label{fig:tmt-framework}
    \vspace{-12pt}
\end{figure}

To address these challenges, we introduce \textbf{TMT}, an inference-time backdoor detector that combines \textbf{T}oken \textbf{M}anifold and latent \textbf{T}ransition modeling using internal representations of the deployed VLA policy (Figure~\ref{fig:tmt-framework}). TMT examines both the structure of input-token representations and how these representations evolve across backbone layers. Our t-SNE visualizations show that the input-token representations of triggered queries are shifted relative to those of benign queries in the projected space, while the two groups still overlap (Figure~\ref{fig:token-motivation}). This motivates examining layer-wise transitions alongside token structure. Trained on benign rollouts, the token manifold branch assesses token structure conditioned on the preceding policy query, while the latent transition branch scores prediction errors in adjacent-layer latent dynamics. Both branches evaluate each policy query, and TMT raises an alarm when either score exceeds its calibrated threshold.

In this paper, we focus on detecting backdoor activation on unseen manipulation tasks, with TMT fitted using benign rollouts from seen tasks. For comparison, we adapt traditional backdoor detectors~\citep{gao2019strip,liu2023detecting,hou2025dede,hu2026dup} and repurpose anomaly and failure detection methods~\citep{ruff2018deep,burda2018exploration,he2024rediffuser,xu2025faildetect,romer2026failure} for detecting backdoor activation (see Appendix~\ref{sec:alternative-detection}). Detectors are fitted using benign rollouts from three LIBERO-10 tasks~\citep{liu2023libero} and evaluated on the remaining tasks under BadVLA~\citep{zhou2026badvla}, GoBA~\citep{zhou2025goal}, and DropVLA~\citep{xu2025dropvla}. In the post-hoc comparison, TMT achieves the highest or joint-highest trigger detection rates (TDRs) of $100\%$, $83.43\%$, and $73.6\%$, respectively, with observed false-rejection rates (FRRs) of $0.57\%$, $8.57\%$, and $0\%$. Branch ablations show that, for GoBA and DropVLA, the token manifold branch provides broader detection coverage on tasks represented during fitting, whereas the latent transition branch provides broader coverage on unseen tasks. We further examine detection in real-world robot execution, and observe that TMT achieves a TDR of $100\%$ with an observed FRR of $0\%$.

Beyond detection, we explore policy purification through self-distillation without a separate clean reference policy. A frozen copy of the backdoored policy serves as the teacher, and the student is initialized from the same checkpoint. We execute the teacher's benign-input actions in both benign and triggered environments, collecting paired observations under a shared action history. The student is trained to reproduce each benign-input action from both observations. Results show that original-task success rates under triggered inputs increase to $92.2$--$95.5\%$, while benign-input success rates change by at most $2.2\%$. Our contributions are threefold:
\begin{itemize}
    \item We introduce TMT, an inference-time VLA backdoor detector that combines input-token structure and latent dynamics learned from benign rollouts.

    \item We propose policy purification through self-distillation without a separate clean reference policy, using the victim's benign-input actions to supervise paired benign and triggered observations.

    \item We adapt existing detection methods to VLA backdoor detection and evaluate them under three kinds of VLA backdoor attacks. Our post-hoc results show strong backdoor detection performance on unseen tasks.
\end{itemize}

\section{Related Work}

\paragraph{Backdoor Attacks on VLA.} Backdoor attacks on VLA policies implant trigger-dependent control behaviors while preserving benign performance. BadVLA~\citep{zhou2026badvla} uses objective-decoupled optimization to separate triggered and benign visual representations and induce control deviations. GoBA~\citep{zhou2025goal} poisons demonstrations with physical-object triggers to redirect high-level goals, whereas DropVLA~\citep{xu2025dropvla} relabels selected action targets to induce low-level failures such as premature gripper opening. Attack demonstrations appear in Appendix~\ref{app:attack-demonstration}. These attacks expose the need for runtime detection across different malicious behaviors. TMT addresses this need by monitoring token manifolds and latent transitions without assuming a specific target action.

\paragraph{Traditional Backdoor Detection.} Existing detectors screen inputs through prediction responses or representation statistics. STRIP~\citep{gao2019strip} measures prediction entropy under perturbations, while TeCo~\citep{liu2023detecting} compares the corruption severity needed to change predicted labels across corruption types. DeDe~\citep{hou2025dede} uses decoder reconstruction errors to screen inputs to self-supervised encoders; DUP-MS~\citep{hu2026dup} combines class-agnostic Mahalanobis distances with spectral statistics of inter-layer feature differences. Concurrent work, TrustVLA~\citep{fu2026trustvla}, uses token- and layer-wise epistemic uncertainty for VLA detection. STRIP, TeCo, DeDe, and DUP-MS were originally developed for image or language models. We adapt them to VLA backdoor detection, but our post-hoc comparison shows limited detection on unseen tasks, whereas TMT achieves higher TDRs.

\paragraph{Anomaly and Failure Detection.} Learning a reference distribution from normal data provides a basis for detecting previously unseen deviations. Deep SVDD~\citep{ruff2018deep} scores learned representations by distance from a hypersphere center, while RND~\citep{burda2018exploration} measures prediction error against fixed random features. ReDiffuser~\citep{he2024rediffuser} uses RND confidence to guide diffusion-based planning. LogPZO~\citep{xu2025faildetect} models successful-rollout observations with a flow and scores their inferred latent noise. FIPER~\citep{romer2026failure} requires both embedding-space RND and action-chunk entropy to exceed calibrated thresholds over short windows. These methods struggle to distinguish benign execution on unseen tasks from trigger-induced execution. Our post-hoc results show that TMT distinguishes these cases more effectively.

\section{Problem Formulation}
\label{sec:problem}

We study inference-time backdoor detection for a potentially backdoored VLA policy $\pi$. At each query $q$, the policy receives an observation $o_q$ and a language instruction $u$, and produces an action chunk $\bm{A}_q=\pi(o_q,u)$. Executing these actions changes the environment and determines subsequent observations, yielding a rollout of $Q$ queries indexed by $q=0,\ldots,Q-1$.
A backdoor trigger may become active at any query and alter subsequent actions and the resulting trajectory.

\paragraph{Defender Capability.} The defender has access to trusted benign rollouts and can inspect the policy's internal representations during inference. Detector fitting and calibration use only benign rollouts, without access to triggered training examples, the trigger specification, or an uncompromised reference policy. Here, benign rollouts denote the absence of a backdoor trigger, regardless of whether the policy successfully completes the task.

\paragraph{Defender Objective.} The defender aims to detect backdoor activation online. Deployment may involve unseen tasks during detector training, requiring cross-task generalization. We aim to identify triggered rollouts with a high detection rate and low trigger-relative detection latency while controlling the rollout-level false-rejection rate (FRR), defined as the fraction of benign rollouts with at least one alarm.

\section{Backdoor Detection}

We introduce TMT, as shown in Figure~\ref{fig:tmt-framework}, a backdoor detector that examines each policy query through two complementary branches. The token manifold branch assesses whether input-token representations conform to the distribution learned from benign queries. The latent transition branch assesses whether observed adjacent-layer transitions agree with those learned from benign inference. Both branches are trained on benign rollouts collected by executing the victim policy without triggers, while the policy remains frozen. After training, we calibrate thresholds using held-out benign rollouts. During online detection, an alarm is raised when either branch's score exceeds its threshold.

\subsection{Motivation}
\label{sec:motivation}

\begin{figure}[t]
    \setlength{\abovecaptionskip}{4pt}
  \centering
  \begin{minipage}[t]{0.32\linewidth}
    \centering
    \includegraphics[width=\linewidth]{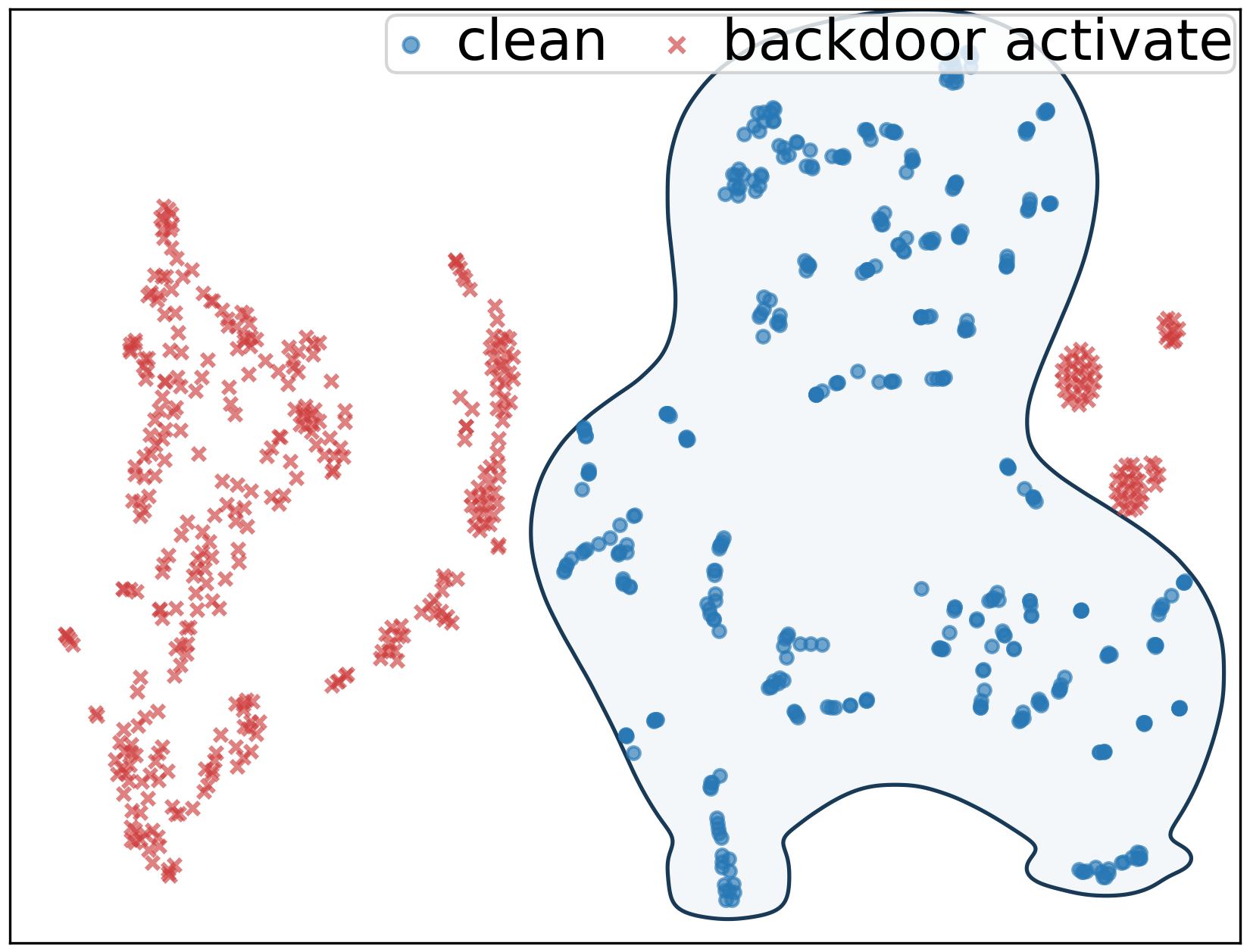}
    \\[0.25em]{\small (a) BadVLA~\citep{zhou2026badvla}}
  \end{minipage}\hfill
  \begin{minipage}[t]{0.32\linewidth}
    \centering
    \includegraphics[width=\linewidth]{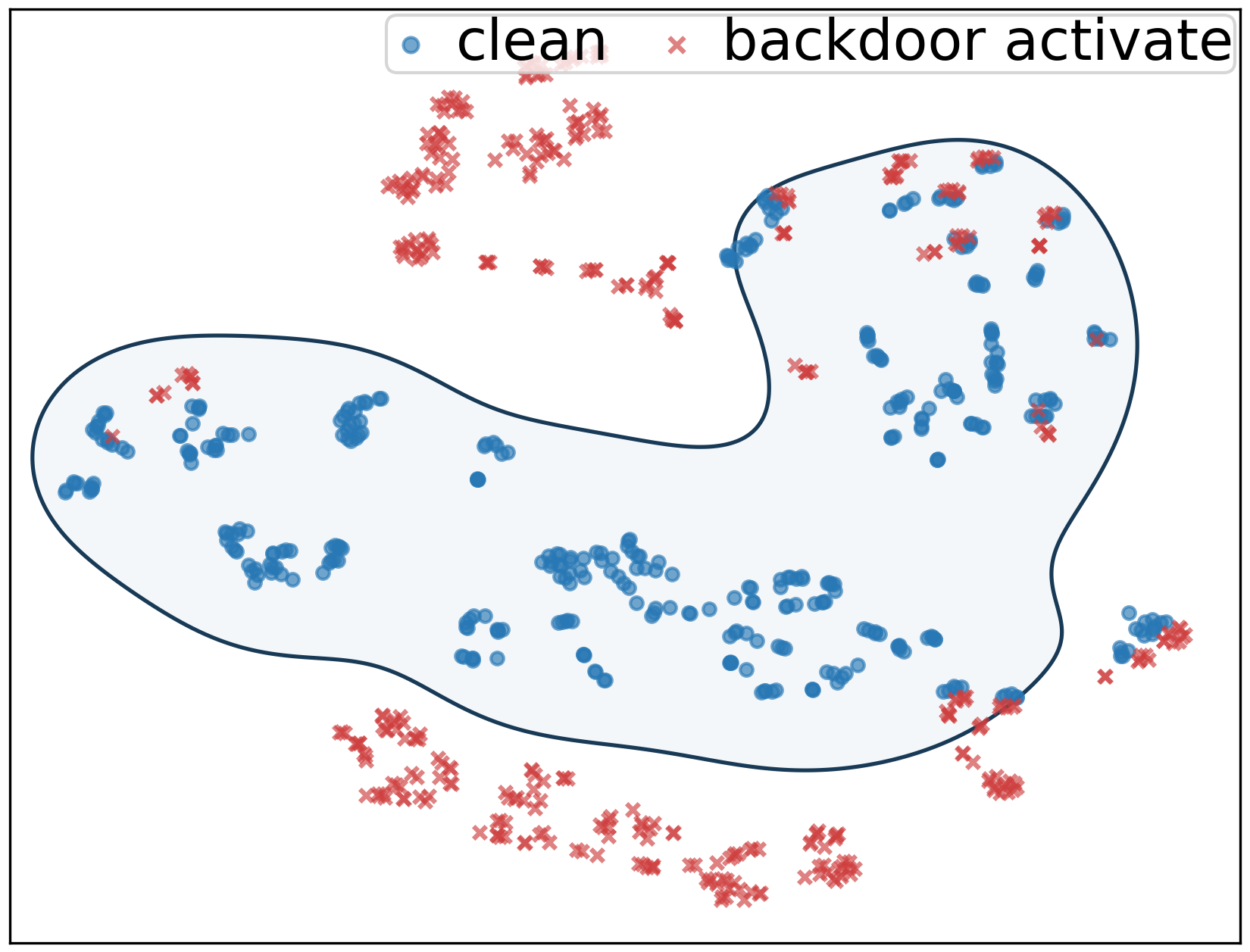}
    \\[0.25em]{\small (b) GoBA~\citep{zhou2025goal}}
  \end{minipage}\hfill
  \begin{minipage}[t]{0.32\linewidth}
    \centering
    \includegraphics[width=\linewidth]{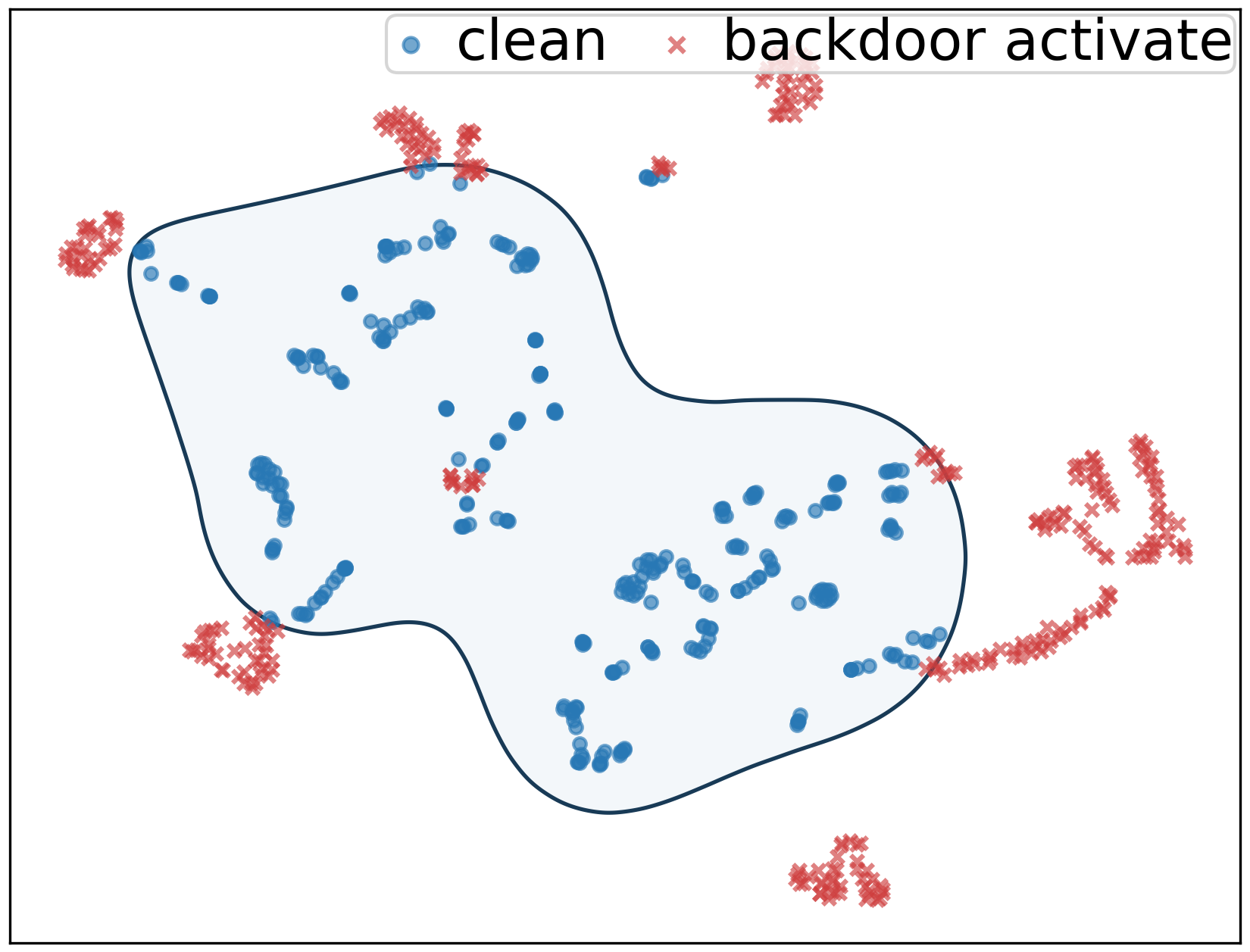}
    \\[0.25em]{\small (c) DropVLA~\citep{xu2025dropvla}}
  \end{minipage}
    \caption{Visualizations of input tokens from backdoored OpenVLA-OFT~\citep{kim2025fine} under BadVLA, GoBA, and DropVLA. Blue circles denote clean queries, and red crosses denote backdoor-active queries.}
  \label{fig:token-motivation}
    \vspace{-12pt}
\end{figure}

\paragraph{Benign Token Representation.}
Existing analyses~\citep{hu2026dup,wang2025backdoor} of backdoor attacks at the input-query level have shown that triggers can induce anomalous representation patterns. With privileged access to the labels of benign and malicious queries, the t-SNE visualization~\citep{van2008visualizing} in Figure~\ref{fig:token-motivation} shows that most benign queries cluster in similar regions. Motivated by these observations, we hypothesize that benign query representations possess a learnable manifold structure and that triggers can shift query representations away from this manifold. We therefore model the benign representation distribution, which we refer to as the benign token manifold, to identify trigger-induced deviations.

\paragraph{Inference Dynamics Consistency.} Just as benign query representations can be characterized by a manifold, their layer-to-layer transitions during inference can also be modeled. \citet{yang2023backdoor} showed that suppressing skip connections in critical layers can reduce backdoor attack success. Inspired by this, we hypothesize that the layer-to-layer transitions of triggered queries deviate from those observed during benign inference. This motivates learning benign latent transitions and examining whether the observed latent dynamics are consistent with these learned relationships.

\subsection{Token Manifold}

Let $\bm{X}\in\mathbb{R}^{N\times d}$ denote the token sequence entering the backbone of the potentially backdoored VLA, where $N$ is the number of valid tokens and $d$ is the embedding dimension. VLA tasks involve sequential interaction, so we use the preceding query to provide temporal context. We condition a velocity prediction network $f_{\bm{\theta}}$, with trainable parameters $\bm{\theta}$, on a history context $\bm{c}\in\mathbb{R}^{d}$, obtained by averaging the valid tokens from the previous policy query: $\bm{c}_q=\operatorname{mean}(\bm{X}_{q-1})$. For the first policy query, we initialize the history context as $\bm{c}_0=\mathbf{0}$.

We sample a noise matrix $\bm{\epsilon}\in\mathbb{R}^{N\times d}$ with elements drawn independently from $\mathcal{N}(0,1)$, and independently sample a transport time $t\sim\operatorname{Uniform}(0,1)$. We construct a linear interpolation between the benign token sequence and Gaussian noise:
\begin{equation}
    \bm{Z}=(1-t)\bm{X}+t\bm{\epsilon}.
    \label{eq:manifold-path}
\end{equation}
We train the velocity prediction network $f_{\bm{\theta}}$ with a flow-matching objective:
\begin{equation}
    \mathcal{L}_{\mathrm{M}}(\bm{\theta})
    =\mathbb{E}_{(\bm{X},\bm{c}),\bm{\epsilon},t}
    \left[
    \frac{\left\lVert
    f_{\bm{\theta}}(\bm{Z},t;\bm{c})-(\bm{\epsilon}-\bm{X})
    \right\rVert_F^2}{Nd}
    \right],
    \label{eq:manifold-loss}
\end{equation}
where $\lVert\cdot\rVert_F$ denotes the Frobenius norm. The expectation is taken over benign queries and their history contexts $(\bm{X},\bm{c})$, independently sampled Gaussian noise $\bm{\epsilon}$, and transport time $t$.

At inference, we evaluate the velocity predictor at $t=0$, where $\bm{Z}=\bm{X}$, to assess the original query representation without noise perturbation. Since $\bm{\epsilon}$ is independent of $(\bm{X},\bm{c})$ and has zero mean, the conditional mean velocity target at this endpoint is $-\bm{X}$. We therefore define the anomaly score as
\begin{equation}
    s_{\mathrm{M}}(\bm{X};\bm{c})
    =\frac{\left\lVert
    \bm{X}+f_{\bm{\theta}}(\bm{X},0;\bm{c})
    \right\rVert_F^2}{Nd}.
    \label{eq:manifold-score}
\end{equation}
This score requires a single forward pass through $f_{\bm{\theta}}$ without noise sampling. Larger scores indicate greater deviation from the learned benign representation patterns and provide evidence of potential backdoor activation.

\subsection{Latent Transition} 

The anomalous tokens can induce the backdoor pattern of the compromised VLA. We model how token representations evolve across backbone layers during benign inference. Let $\bm{H}_\ell\in\mathbb{R}^{N\times d}$ denote the token sequence at backbone depth $\ell=0,\ldots,L-1$, with $\bm{H}_0=\bm{X}$. We define the latent transition between adjacent layers as
\begin{equation}
    \bm{\Delta}_\ell
    =\frac{\bm{H}_{\ell+1}-\bm{H}_\ell}{\sigma_\ell},
    \label{eq:latent-transition}
\end{equation}
where $\sigma_\ell>0$ is a fixed layer-specific scale estimated from benign training data to normalize differences in transition magnitude across layers. We train a latent transition predictor $g_{\bm{\phi}}$ with parameters $\bm{\phi}$ to estimate $\bm{\Delta}_\ell$ from $\bm{H}_\ell$:
\begin{equation}
    \mathcal{L}_{\mathrm{D}}(\bm{\phi})=
    \mathbb{E}_{\bm{X},\ell}
    \left[
    \frac{\left\lVert g_{\bm{\phi}}(\bm{H}_\ell,\ell)-\bm{\Delta}_\ell\right\rVert_F^2}{Nd}
    \right],
\label{eq:transition-loss}
\end{equation}
where the expectation is taken over clean queries $\bm{X}$ and the layer index $\ell$, sampled independently and uniformly from the adjacent-layer transitions.

\paragraph{Transition Selection.} We fit the detector using verified benign rollouts and initialize transition selection on tasks represented in this training set. The token manifold branch flags suspicious rollouts during subsequent executions of these tasks. Once a flagged rollout is confirmed as a backdoor incident, we compare its prediction errors across all adjacent-layer transitions with the corresponding benign references. This confirmation uses only detector scores and benign reference statistics and does not require an externally supplied malicious label (see Appendix~\ref{app:Detail of Transition Selection}). Among transitions exhibiting significant deviations, we select the one with the largest standardized excess and denote its index by $\ell^\star$, as shown in Figure~\ref{fig:tmt-framework}. The selected transition remains fixed during subsequent online monitoring, including rollouts from unseen tasks. Both branches remain active, and either branch can independently raise an alarm.

For each subsequent policy query, we compute the consistency score at the selected transition by averaging the $K$ largest token prediction errors:
\begin{equation}
    s_{\mathrm{D}}(\bm{H}_{\ell^\star},\ell^\star)
    =
    \frac{1}{Kd}\sum_{i\in\mathcal{I}_K}
    \left\lVert
    [g_{\bm{\phi}}(\bm{H}_{\ell^\star},\ell^\star)
    -\bm{\Delta}_{\ell^\star}]_i
    \right\rVert_2^2.
    \label{eq:transition-token-error}
\end{equation}
Here, $\mathcal{I}_K$ contains the indices of the $K$ tokens with the largest prediction errors, and $[\cdot]_i$ denotes the $i$-th row. We normalize this score by a fixed error scale estimated from benign queries at the selected transition. Larger normalized scores indicate greater deviation from benign latent dynamics.

\subsection{Calibration and Online Detection}
\label{sec:threshold-calibration}

After training $f_{\bm{\theta}}$ and $g_{\bm{\phi}}$, we fix these two networks and calibrate the branch-specific thresholds to enable online detection. After fitting the detection network and selecting the monitored transitions, We then use the retained benign queries to calibrate branch-specific detection thresholds for online monitoring.

For the token manifold branch, we set the threshold to the empirical $0.99$ quantile of the benign rollout maximum scores plus a margin proportional to their median absolute deviation (MAD)~\citep{rousseeuw1993alternatives}:
\begin{equation}
    \tau_{\mathrm{M}} =
    \operatorname{Quantile}_{0.99}\!\left[
    \max_q s_{\mathrm{M}}(\bm{X}_q;\bm{c}_q)
    \right]
    +\operatorname{MAD}\!\left[
    \max_q s_{\mathrm{M}}(\bm{X}_q;\bm{c}_q)
    \right].
    \label{eq:manifold-threshold}
\end{equation}

For the latent transition branch, we use the maximum scores from held-out benign rollouts to compute the empirical tail probability of the current online score. Its negative logarithm gives the calibrated detection score. We set $\tau_{\mathrm{D}}$ to the negative logarithm of a prescribed tail-probability cutoff.

\paragraph{Online Detection.} In subsequent rollouts, the token manifold branch evaluates $s_{\mathrm{M}}(\bm{X};\bm{c})$ at each query, while the latent transition branch monitors the selected layer $\ell^\star$. Let $\tilde{s}_{\mathrm{D}}$ denote its calibrated detection score defined above. An alarm is raised when either score exceeds its threshold:
\begin{equation}
    b =
    \mathbf{1}\!\left[
    s_{\mathrm{M}}(\bm{X};\bm{c})>\tau_{\mathrm{M}}
    \ \lor\
    s_{\mathrm{D}}(\bm{H}_{\ell^\star},\ell^\star)
    >\tau_{\mathrm{D}}
    \right].
    \label{eq:joint-alarm}
\end{equation}
Here, $b=1$ indicates a detection alarm.

\section{Backdoor Purification}
\label{sec:purification}

To suppress backdoor-induced behavior while preserving benign task performance, we propose a purification method based on policy self-distillation with teacher-driven paired rollouts. Here, the teacher is a frozen copy of the original backdoored policy $\pi$, and the student $\pi_{\bm{\psi}}$ is initialized from the same checkpoint, with trainable parameters $\bm{\psi}$. The teacher's actions on benign observations provide supervision for preserving benign behavior and correcting actions under malicious observations. 

\paragraph{Teacher-driven Paired Rollouts.} At each policy query, the teacher predicts an action chunk from the benign observation. We execute the same actions in both the benign and malicious environments, collecting paired observations under a shared action history. Let $o$ and $\tilde{o}$ denote the benign and malicious observations at corresponding queries, with the same language instruction $u$. The teacher's action chunk $\pi(o,u)$ advances both environments and provides the supervision target for both observations.

\paragraph{Action Distillation.} We train the student to reproduce the teacher's benign-input actions from both observations:
\begin{equation}
\mathcal{L}_{\mathrm{P}}(\bm{\psi})
=
\mathbb{E}_{(o,\tilde{o},u)}\!\left[
\left\lVert\pi_{\bm{\psi}}(o,u)-\pi(o,u)\right\rVert_F^2
+
\left\lVert\pi_{\bm{\psi}}(\tilde{o},u)-\pi(o,u)\right\rVert_F^2
\right].
\label{eq:purification-loss}
\end{equation}
The expectation is taken over paired observations and their shared language instructions from the collected rollouts. The first term preserves the teacher's behavior on benign inputs, while the second trains the student to reproduce the corresponding benign actions in the
presence of a trigger. After training, the student is used throughout each evaluation rollout.

\section{Experiments}

\subsection{Experimental Setup}

\paragraph{Attacks.} We consider three attacks: BadVLA~\citep{zhou2026badvla}, GoBA~\citep{zhou2025goal}, and DropVLA~\citep{xu2025dropvla}.
In our experiments, BadVLA uses a white patch as the trigger and aims to cause task failure. GoBA targets high-level goals, using a physical ``toxic box'' as the trigger to redirect the VLA policy toward an attacker-specified goal. DropVLA targets low-level actions using the textual trigger ``carefully'' together with a digital red-dot trigger. When both triggers are present, it induces the VLA policy to open the gripper, causing task failure. Details of the attack performance can be found in Appendix~\ref{app:purification}, and demonstrations can be found in Appendix~\ref{app:attack-demonstration}.

\paragraph{Settings.} For simulation experiments, we evaluate backdoor detection on OpenVLA-OFT~\citep{kim2025fine}, VLA-Adapter~\citep{wang2026vla}, and $\pi_{0.5}$~\citep{intelligence2025pi_} using the LIBERO benchmark~\citep{liu2023libero}. For real-world experiments, we deploy $\pi_{0.5}$ on a Trossen Robotics WidowX AI arm~\footnote{www.trossenrobotics.com} and collect data for four tasks. Real-world demonstrations are shown in Figure~\ref{fig:real-world-attack-demo} in Appendix~\ref{app:real-world-attack-demonstrations}.  

\paragraph{Baselines.} We compare against three groups of baselines. Traditional backdoor detectors include STRIP~\citep{gao2019strip},
TeCo~\citep{liu2023detecting}, DeDe~\citep{hou2025dede}, and DUP-MS~\citep{hu2026dup}. Anomaly and failure detection methods include Deep SVDD~\citep{ruff2018deep}, RND~\citep{burda2018exploration}, ReDiffuser~\citep{he2024rediffuser}, LogpZO~\citep{xu2025faildetect}, and FIPER~\citep{romer2026failure}. We also compare with concurrent work TrustVLA~\citep{fu2026trustvla}, which specifically targets inference-time backdoor detection in VLA models. All details related to the VLA implementation can be found in Appendix~\ref{sec:alternative-detection}.

\paragraph{Detector Architectures.} The token manifold velocity predictor $f_{\bm{\theta}}$ uses four residual MLP blocks. The latent transition predictor $g_{\bm{\phi}}$ uses a two-layer Transformer encoder with eight attention heads per layer.

\paragraph{Evaluation Metrics.} We evaluate detection at the rollout level. A rollout is flagged if an alarm occurs at any policy query. The trigger detection rate (TDR) and false-rejection rate (FRR) are defined as
\begin{equation}
    \mathrm{TDR} = \frac{N_{\mathrm{TP}}}{N_{\mathrm{malicious}}},
    \qquad
    \mathrm{FRR} = \frac{N_{\mathrm{FP}}}{N_{\mathrm{benign}}},
\end{equation}
where $N_{\mathrm{TP}}$ and $N_{\mathrm{FP}}$ count flagged malicious and benign rollouts, respectively, and $N_{\mathrm{malicious}}$ and $N_{\mathrm{benign}}$ denote their respective totals. Higher TDR and lower FRR indicate better detection performance. For threshold-independent evaluation, we report the area under the ROC curve (AUC)~\citep{fawcett2006introduction}. Higher AUC indicates better separation between malicious and benign rollouts. Detection latency is the difference in policy-query indices between trigger onset and the first alarm at or after onset, averaged over malicious rollouts with such an alarm. Lower latency indicates faster detection.

\subsection{Main Experimental Results}

We evaluate backdoor detection on tasks unseen during detector fitting using backdoored OpenVLA-OFT~\citep{kim2025fine} on LIBERO-10~\citep{liu2023libero}, a suite of ten long-horizon manipulation tasks. The detectors are trained on benign backdoored VLA inference-time rollouts randomly selected from three of the tasks, while the remaining seven tasks form the evaluation task pool. This setup is designed to test whether the detector can identify backdoor activations that fall outside the range of tasks available during training.

\begin{table}[t]
    \setlength{\abovecaptionskip}{0pt}
    \setlength{\belowcaptionskip}{4pt}
    \centering
    \caption{Post-hoc comparison on tasks unseen during detector fitting, using backdoored OpenVLA-OFT. We report rollout-level AUC, TDR, FRR, and mean detection latency in policy queries (Query), measured from trigger onset to the first alarm at or after onset. Latency is averaged over malicious rollouts with such an alarm; -- indicates that none occur. Bold and underlining indicate the highest and second-highest distinct values, respectively, for AUC and TDR.}
    \label{tab:unseen-main}

    \setlength{\tabcolsep}{2.5pt}
    \renewcommand{\arraystretch}{1.15}
    \footnotesize
    
    \resizebox{\linewidth}{!}{%
    \begin{tabular}{@{}l*{3}{rrrr}@{}}
    \toprule
    & \multicolumn{4}{c}{BadVLA}
    & \multicolumn{4}{c}{GoBA}
    & \multicolumn{4}{c}{DropVLA} \\
    \cmidrule(lr){2-5}
    \cmidrule(lr){6-9}
    \cmidrule(lr){10-13}
    
        Method
        & AUC$\uparrow$ & TDR$\uparrow$
        & FRR$\downarrow$ & Query$\downarrow$
        & AUC$\uparrow$ & TDR$\uparrow$
        & FRR$\downarrow$ & Query$\downarrow$
        & AUC$\uparrow$ & TDR$\uparrow$
        & FRR$\downarrow$ & Query$\downarrow$ \\
    \midrule
    
    \rowcolor{gray!12}
    \multicolumn{13}{@{}l}{\textit{Backdoor Detection}} \\
        
        STRIP
        & \underline{0.770} & \underline{22.57} & 0.00 & 13.44
        & \underline{0.940} & 4.00 & 0.00 & 14.00
        & 0.749 & 0.80 & 2.00 & 30.50 \\
        
        TeCo
        & 0.251 & 0.57 & 7.71 & 48.00
        & 0.662 & 5.71 & 0.86 & 2.65
        & 0.673 & 6.00 & 1.60 & 24.27 \\
        
        DeDe
        & 0.535 & 0.57 & 3.43 & 30.00
        & 0.631 & 2.00 & 6.86 & 36.14
        & 0.367 & 0.00 & 0.00 & -- \\
        
        DUP-MS
        & 0.001 & 0.00 & 0.29 & --
        & 0.841 & 4.86 & 0.00 & 13.71
        & 0.570 & 1.60 & 2.00 & 39.00 \\
        
        TrustVLA
        & \textbf{1.000} & \textbf{100.00} & 0.57 & 0.00
        & 0.268 & 18.00 & 50.86 & 12.06
        & 0.475 & 0.40 & 0.40 & -- \\
        
        \rowcolor{gray!12}
        \multicolumn{13}{@{}l}{
        \textit{Anomaly and Failure Detection}} \\
        
        Deep SVDD
        & 0.033 & 0.00 & 43.71 & --
        & 0.336 & \underline{42.86} & 42.86 & 0.00
        & 0.638 & 20.00 & 16.80 & 0.00 \\
        
        RND
        & 0.448 & 1.43 & 8.57 & 42.00
        & 0.189 & 0.57 & 37.43 & 62.00
        & 0.592 & 16.80 & 30.40 & 33.48 \\
        
        ReDiffuser
        & \textbf{1.000} & \textbf{100.00} & 4.00 & 0.00
        & 0.626 & 15.14 & 0.29 & 13.45
        & 0.716 & 22.80 & 0.00 & 16.42 \\
        
        LogpZO
        & 0.503 & 0.00 & 2.86 & --
        & 0.138 & 0.29 & 41.43 & 60.00
        & 0.315 & 10.00 & 40.00 & 33.24 \\
        
        FIPER
        & 0.000 & 0.00 & 1.71 & --
        & 0.297 & 5.14 & 1.14 & 51.06
        & \underline{0.955} & \underline{26.00} & 0.00 & 49.00 \\
        
        \rowcolor{gray!12}
        \multicolumn{13}{@{}l}{\textit{Our Method}} \\
        
        \textbf{TMT}
        & \textbf{1.000} & \textbf{100.00} & 0.57 & 0.00
        & \textbf{0.983} & \textbf{83.43} & 8.57 & 0.28
        & \textbf{1.000} & \textbf{73.60} & 0.00 & 5.68 \\
        
    \bottomrule
    \end{tabular}%
    }

    \vspace{-12pt}
\end{table}

In this reanalysis, as shown in Table~\ref{tab:unseen-main}, TMT achieves the highest or joint-highest AUC and TDR across the three attacks. For BadVLA, an untargeted attack that disrupts task execution, TMT achieves a TDR of $100\%$ with an observed FRR of $0.57\%$. TrustVLA and ReDiffuser also achieve $100\%$ TDR, with FRRs of $0.57\%$ and 4\%, respectively.

GoBA presents a different detection challenge: it redirects the policy toward an attacker-specified goal through a sequence of actions that can individually resemble benign behavior. Local actions such as approaching an object or opening the gripper can occur in both benign and malicious rollouts, while their role in the overall execution differs. TMT achieves an AUC of $98.3\%$ and a TDR of $83.43\%$, with a FRR of $8.57\%$. Among the traditional backdoor detectors, STRIP and DUP-MS achieve AUCs of $94\%$ and $84.1\%$ but detect only $4\%$ and $4.86\%$ of malicious rollouts, respectively, at the calibrated thresholds. The AUCs of the anomaly and failure detection baselines range from $13.8\%$ to $62.6\%$. ReDiffuser's AUC and TDR drop from $100\%$ and $100\%$ on BadVLA to $62.6\%$ and $15.14\%$ on GoBA. Unlike the disrupted execution observed under BadVLA, GoBA can produce coherent, goal-directed behavior. This difference may help explain why detectors of execution anomalies struggle to distinguish its malicious rollouts from benign ones.

DropVLA targets low-level control by inducing gripper-opening actions when triggers are present. This behavior interferes with grasping and can cause repeated unsuccessful grasp attempts. We observe that TMT detects $73.6\%$ of malicious rollouts without rejecting any benign rollout in the evaluated set. By comparison, FIPER achieves an AUC of $95.5\%$ and the second-highest TDR of $26\%$ at the same observed FRR of $0\%$. Deep SVDD detects $20\%$ of malicious rollouts, accompanied by an FRR of $16.8\%$. 

Additional results on $\pi_{0.5}$~\citep{intelligence2025pi_} and VLA-Adapter~\citep{wang2026vla} show that TMT achieves the highest TDR on both victim models (Appendix~\ref{sec:cross-victim}). We further consider a privileged all-seen setting in which benign rollouts from all eligible tasks are available for detector fitting (Appendix~\ref{app:all-seen}). In this post-hoc comparison, several existing backdoor, anomaly, and failure detectors also achieve high detection rates, while TMT attains the highest or joint-highest AUC across all three attacks.

\begin{figure}[b]
    \centering
    \includegraphics[width=0.49\linewidth]
        {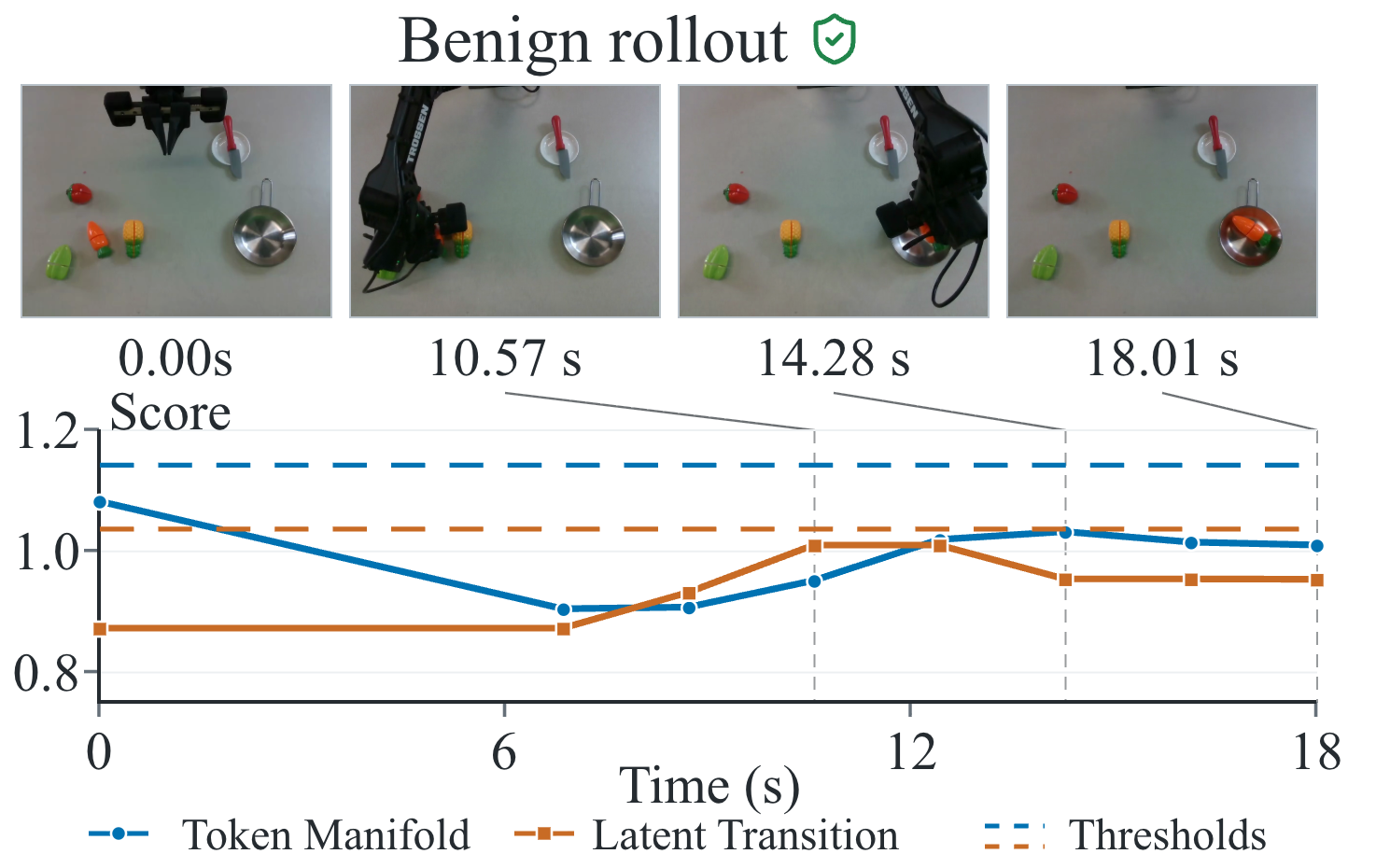}
    \hfill
    \includegraphics[width=0.49\linewidth]
        {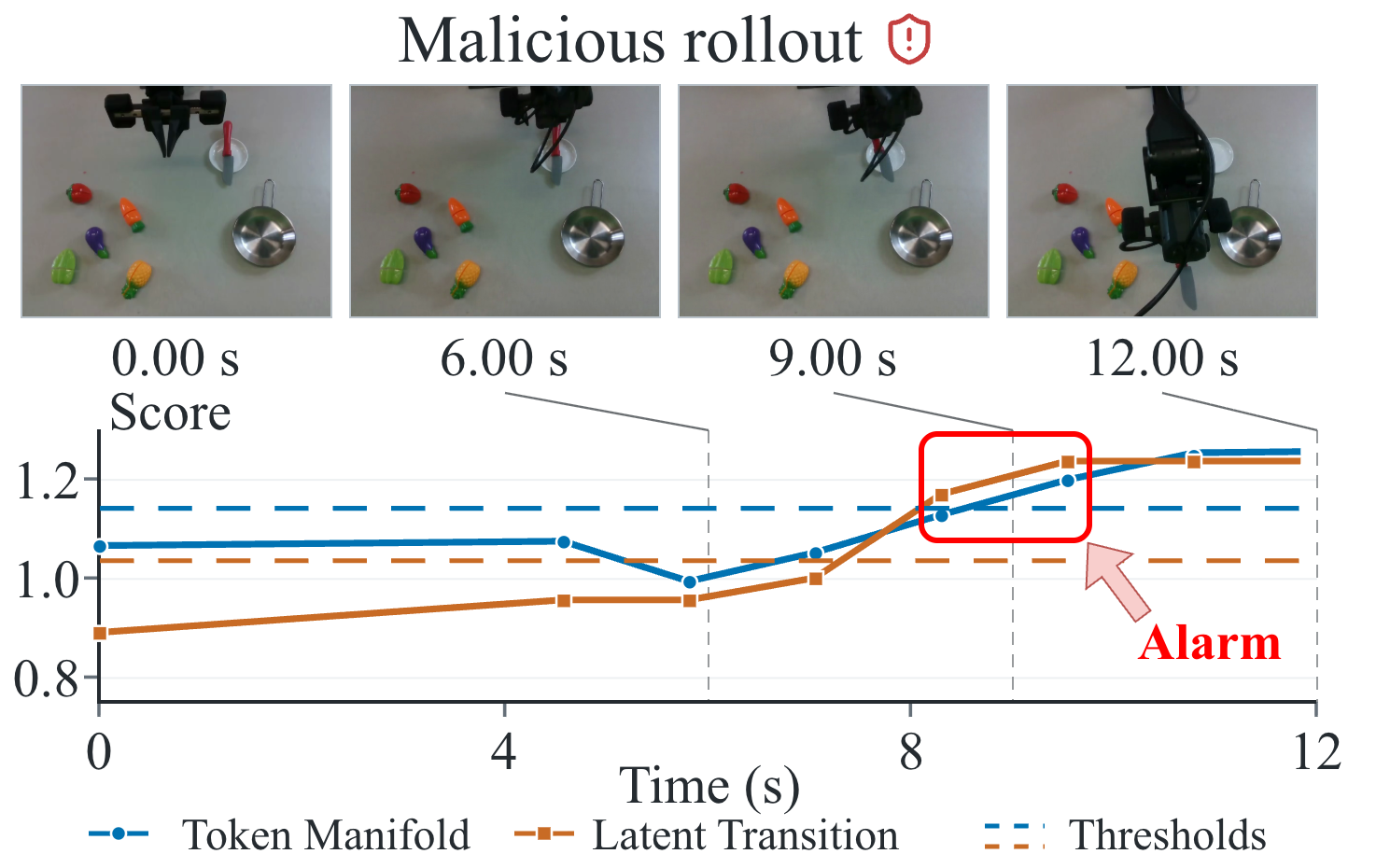}
    \caption{Benign and GoBA-triggered real-world behavior with detection-score traces. Dashed lines denote the
    calibrated thresholds of the token manifold and latent transition branches. The red box highlights threshold exceedances that trigger an alarm. Frames are drawn from separate demonstration recordings.}
    \label{fig:real-world-detection}
\end{figure}

\subsection{Real-world Experiments}
\label{sec:real-world}

We evaluate real-world backdoor detection against GoBA on four pick-and-place tasks using $\pi_{0.5}$ deployed on a Trossen Robotics WidowX AI arm (see Appendix~\ref{app:real-world-attack-demonstrations} for attack demonstrations). We train the detector using benign rollouts from two randomly selected tasks and evaluate benign and GoBA-triggered rollouts on the other two tasks, with 30 rollouts per evaluation task. 

\begingroup
\setlength{\intextsep}{4pt}
\setlength{\columnsep}{12pt}
\begin{wraptable}[16]{r}{0.44\textwidth}
    \centering
\setlength{\abovecaptionskip}{0pt}
\setlength{\belowcaptionskip}{4pt}
\caption{Real-world GoBA detection on unseen tasks.}
\label{tab:real-world}

    \footnotesize
    \setlength{\tabcolsep}{2pt}
    \renewcommand{\arraystretch}{1.03}
    \sisetup{
        mode = text,
        detect-weight = true,
        detect-family = true
    }

    \begin{tabular*}{\linewidth}{
        @{\extracolsep{\fill}}
        l
        S[table-format=1.3]
        S[table-format=3.2]
        S[table-format=2.2]
        @{}
    }
        \toprule
        Method
        & {AUC$\uparrow$}
        & {TDR$\uparrow$}
        & {FRR$\downarrow$} \\
        \midrule

        \rowcolor{gray!12}[0pt][0pt]
        \multicolumn{4}{@{}l@{}}{
            \textit{Backdoor Detection}} \\

        DeDe
        & 0.833 & 0.00 & 10.00 \\
        DUP-MS
        & \bfseries 1.000
        & \bfseries 100.00
        & 66.67 \\
        TrustVLA
        & 0.528 & 0.00 & 8.33 \\

        \rowcolor{gray!12}[0pt][0pt]
        \multicolumn{4}{@{}l@{}}{
            \textit{Anomaly and Failure Detection}} \\

        Deep SVDD
        & 0.917 & 83.33 & 16.67 \\
        RND
        & \bfseries 1.000
        & \bfseries 100.00
        & \bfseries 0.00 \\
        ReDiffuser
        & \bfseries 1.000
        & \bfseries 100.00
        & 16.67 \\
        LogpZO
        & \bfseries 1.000
        & 66.67
        & \bfseries 0.00 \\
        FIPER
        & 0.958
        & \bfseries 100.00
        & 26.67 \\

        \rowcolor{gray!12}[0pt][0pt]
        \multicolumn{4}{@{}l@{}}{
            \textit{Our Method}} \\

        \textbf{TMT}
        & \bfseries 1.000
        & \bfseries 100.00
        & \bfseries 0.00 \\
        \bottomrule
    \end{tabular*}

\end{wraptable}

Table~\ref{tab:real-world} shows that TMT achieves an AUC of $100\%$ and detects all evaluated malicious rollouts without rejecting any benign rollout. RND matches TMT on all three metrics, while DUP-MS, ReDiffuser, and FIPER also achieve $100\%$ TDR but incur higher FRRs. Figure~\ref{fig:real-world-detection} illustrates benign and GoBA-triggered behavior alongside example detection-score traces. Both branches remain below their respective thresholds in the benign example, whereas the malicious trace exhibits threshold exceedances that activate the alarm.

Real-time deployment also constrains the number of policy evaluations available for each observation. STRIP~\citep{gao2019strip} evaluates predictions across multiple perturbed inputs, while TeCo~\citep{liu2023detecting} probes prediction changes across corruption types and severity levels. The repeated policy evaluations required by their VLA adaptations exceed the per-query time budget of our robotic setup, so we exclude them from the real-time comparison.

\par\endgroup

\subsection{Ablation and Mechanism Analysis}

\begingroup
\setlength{\intextsep}{4pt}
\setlength{\columnsep}{12pt}

\begin{wraptable}{r}{0.44\textwidth}
\centering
\setlength{\abovecaptionskip}{0pt}
\setlength{\belowcaptionskip}{4pt}
\caption{
DropVLA branch ablation on OpenVLA-OFT (post-hoc).
M: token manifold branch; L: latent transition branch.
}
\label{tab:branch-ablation}

\begingroup
\setlength{\tabcolsep}{3pt}
\renewcommand{\arraystretch}{1.10}
\footnotesize

\begin{tabular*}{\linewidth}{@{\extracolsep{\fill}}lrrr@{}}
\toprule
Branch & TDR$\uparrow$ & FRR$\downarrow$ & Query$\downarrow$ \\
\midrule

\rowcolor{gray!12}[0pt][0pt]
\multicolumn{4}{@{}l@{}}{\textit{All-seen}} \\

M only
& \textbf{100.00} & 1.25 & 9.45 \\

L only
& 72.00 & 0.00 & 3.66 \\

\textbf{TMT (M+L)}
& \textbf{100.00} & 1.25 & 2.43 \\
\midrule

\rowcolor{gray!12}[0pt][0pt]
\multicolumn{4}{@{}l@{}}{\textit{Unseen}} \\

M only
& 15.20 & 0.00 & 15.47 \\

L only
& \textbf{73.60} & 0.00 & 6.25 \\

\textbf{TMT (M+L)}
& \textbf{73.60} & 0.00 & 5.68 \\

\bottomrule
\end{tabular*}
\endgroup

\end{wraptable}

The token manifold branch identifies suspicious rollouts through deviations from benign input-token structure. The latent transition branch uses this incident evidence to select an adjacent-layer transition for subsequent monitoring. Table~\ref{tab:branch-ablation} presents the DropVLA ablation; corresponding BadVLA and GoBA results appear in Appendix~\ref{app:branch-ablation}.

\paragraph{Detection Coverage.} The relative contributions of the two branches vary with task availability. Under all-seen DropVLA, the token manifold branch detects all malicious rollouts, while the latent transition branch achieves a TDR of $72\%$. On unseen tasks, the latent transition branch achieves a TDR of $73.6\%$, compared with $15.2\%$ for the token manifold branch. Thus, the token manifold branch provides broader detection coverage when the evaluation tasks are represented during fitting, whereas the latent transition branch becomes
the main source of detection on unseen tasks. The other attacks provide similar evidence
(Appendix~\ref{app:branch-ablation}).

\paragraph{Detection Latency.} The latent transition branch enables earlier alarms when combined with the token manifold branch. On all-seen DropVLA, combining the branches reduces mean latency from $9.45$ to $2.43$ queries while retaining $100\%$ TDR and $1.25\%$ FRR. On unseen DropVLA, joint detection retains the coverage of the latent transition branch and achieves a mean latency of $5.68$ queries. GoBA also shows reduced latency, while BadVLA alarms occur at trigger onset (Appendix~\ref{app:branch-ablation}).

\par\endgroup

\subsection{Backdoor Purification}
\label{sec:purification-results}

We evaluate policy self-distillation on LIBERO-10 by comparing the original backdoored policies with their purified students. In this preliminary evaluation, original-task success rates under attacks increase from $0\%$, $0.4\%$, and $0\%$ to $92.2\%$, $94.8\%$, and $95.5\%$ for BadVLA, GoBA, and DropVLA, respectively, while benign-input success rates change by $-2.2\%$, $-0.4\%$, and $+0.4\%$. Appendix~\ref{app:purification} provides paired-rollout demonstrations and the complete results in Table~\ref{tab:purification}.

\section{Conclusion}

We introduced TMT for runtime backdoor detection in VLA policies on tasks unseen during detector fitting. TMT models benign input-token manifold and adjacent-layer latent transitions on a frozen backbone. The post-hoc comparison under three popular backdoor attacks across three different victim models shows that these signals enable detection across different forms of backdoor activation. Branch ablations reveal broader token-manifold coverage on familiar tasks and a larger role for latent transitions on unseen tasks. We also explored policy self-distillation using paired benign and malicious observations as a purification extension. These results motivate examining both input representations and internal inference dynamics when monitoring VLA policies for backdoor activation.

\section*{AI Use Statement}

We used generative AI tools solely for language polishing and limited assistance with code generation. All AI-assisted content was reviewed and verified by the authors, and we take full responsibility for the final manuscript and all associated claims, code, and artifacts.

\section*{Reproducibility Statement}

To support reproducibility, we provide our implementation code in the supplementary materials, together with demonstration videos of the real-world experiments. 

\bibliography{iclr2027_conference}
\bibliographystyle{iclr2027_conference}

\newpage
\appendix
\section*{Appendix}

\section{Baseline Adaptations for VLA Backdoor Detection}
\label{sec:alternative-detection}

\subsection{Anomaly and Failure Detection as Backdoor Detection}
\label{app:anomaly-as-backdoor-detection}

Backdoor activation can introduce deviations from benign execution through changes in visual observations, policy representations, robot states, or their temporal evolution. Relative to a reference distribution learned from trusted benign rollouts, these deviations can be viewed as out-of-distribution (OOD) behavior. This motivates using anomaly and failure detection methods for backdoor detection: they learn patterns of benign execution and assess departures from those patterns without requiring examples of the specific attack.

The connection is particularly relevant when backdoor activation disrupts task execution. For example, repeated grasp failures or unexpected gripper opening can produce observation and state sequences that differ from benign rollouts. Methods designed to detect execution anomalies or predict task failure may therefore provide useful signals for identifying backdoor activation.

Goal-directed attacks pose a different challenge. A malicious rollout can consist of locally plausible actions while pursuing an object or goal that conflicts with the user's instruction. Meanwhile, unfamiliar tasks can introduce benign distribution shifts. We therefore evaluate whether anomaly and failure detection scores distinguish attack-induced deviations from legitimate variation, particularly on tasks unseen during detector fitting.

\subsection{Baseline Implementation and Adaptation}
\label{app:baseline-implementation}

We adapt the baselines to produce a detection score at each VLA policy query. The available inputs include visual observations, robot states, policy representations, and query history, depending on the method. The victim policy remains frozen during detector fitting and evaluation. Thresholded decisions and detection latency follow the rollout-level definitions in the main text.

\paragraph{Backdoor Detection Baselines.} STRIP~\citep{gao2019strip} and TeCo~\citep{liu2023detecting} assess policy responses to perturbed observations. DeDe~\citep{hou2025dede} uses reconstruction discrepancies, while DUP-MS~\citep{hu2026dup} uses representation statistics. TrustVLA~\citep{fu2026trustvla} provides a VLA-specific comparison based on internal representation uncertainty. These methods use the VLA's available predictions or representations to produce query-level detection scores.

\paragraph{Anomaly and Failure Detection Baselines.} The RND~\citep{burda2018exploration} baseline uses robot-state inputs. ReDiffuser~\citep{he2024rediffuser} uses its RND confidence module, and FIPER~\citep{romer2026failure} uses the RND-OE detector component. Deep SVDD~\citep{ruff2018deep} scores the distance from its learned benign representation center. LogpZO~\citep{xu2025faildetect} applies a flow-based scorer to observation representations supplied by a fixed visual encoder together with robot state. These adaptations assess deviations from benign execution using their respective input representations.

\section{Privileged Access to All Tasks}
\label{app:all-seen}

We consider a privileged setting in which detectors are fitted using benign rollouts from all tasks. Table~\ref{tab:all-seen} reports the post-hoc comparison on backdoored OpenVLA-OFT~\citep{kim2025fine} under the same three attacks.

For BadVLA~\citep{zhou2026badvla}, TMT and ReDiffuser both achieve 100\% TDR, with observed FRRs of 1.6\% and 3.6\%, respectively, while TrustVLA reaches the same TDR with 0.2\% FRR. For GoBA~\citep{zhou2025goal}, TMT achieves 95.4\% TDR with 0.6\% FRR. Although DUP-MS and LogpZO have higher TDRs of 99.2\% and 99.6\%, their observed FRRs are 62.8\% and 11.6\%, respectively. For DropVLA~\citep{xu2025dropvla}, TMT and FIPER both detect all malicious rollouts, with observed FRRs of 1.25\% and 6.88\%. These comparisons show that access to familiar tasks does not remove the need to assess detection coverage jointly with false rejections.

\begin{table}[t]
    \centering
\caption{
    Post-hoc backdoor detection with access to benign rollouts from all eligible tasks, using backdoored OpenVLA-OFT. TDR and observed FRR are percentages. Query reports detection latency in policy queries; -- indicates no detections.
    }
\label{tab:all-seen}

    \begingroup
    \setlength{\tabcolsep}{2.5pt}
    \renewcommand{\arraystretch}{1.15}
    \footnotesize
    
    \resizebox{\linewidth}{!}{%
    \begin{tabular}{@{}l*{3}{rrrr}@{}}
    \toprule
    & \multicolumn{4}{c}{BadVLA}
    & \multicolumn{4}{c}{GoBA}
    & \multicolumn{4}{c}{DropVLA} \\
    \cmidrule(lr){2-5}
    \cmidrule(lr){6-9}
    \cmidrule(lr){10-13}
    
    Method
    & AUC$\uparrow$ & TDR$\uparrow$
    & FRR$\downarrow$ & Query$\downarrow$
    & AUC$\uparrow$ & TDR$\uparrow$
    & FRR$\downarrow$ & Query$\downarrow$
    & AUC$\uparrow$ & TDR$\uparrow$
    & FRR$\downarrow$ & Query$\downarrow$ \\
    \midrule
    
    \rowcolor{gray!12}
    \multicolumn{13}{@{}l}{\textit{Backdoor Detection}} \\
    
    STRIP
    & 0.482 & 30.00 & 3.00 & 20.64
    & 0.949 & 80.40 & 3.00 & 1.56
    & 0.775 & 22.50 & 3.75 & 15.50 \\
    
    TeCo
    & 0.243 & 14.60 & 64.00 & 32.36
    & 0.612 & 68.00 & 49.80 & 4.46
    & 0.641 & 63.13 & 41.25 & 15.63 \\
    
    DeDe
    & 0.958 & 83.40 & 4.40 & 10.89
    & 0.906 & 66.80 & 5.00 & 10.31
    & 0.847 & 36.88 & 3.75 & 23.08 \\
    
    DUP-MS
    & 0.000 & 0.00 & 63.80 & --
    & 0.932 & \underline{99.20} & 62.80 & 6.91
    & 0.651 & 69.38 & 61.88 & 11.24 \\
    
    TrustVLA
    & \textbf{1.000} & \textbf{100.00} & 0.20 & 0.00
    & 0.325 & 18.60 & 22.20 & 9.90
    & 0.441 & 10.00 & 21.88 & 29.94 \\
    
    \rowcolor{gray!12}
    \multicolumn{13}{@{}l}{\textit{Anomaly and Failure Detection}} \\
    
    Deep SVDD
    & 0.725 & 33.40 & 3.60 & 3.08
    & 0.731 & 61.80 & 4.00 & 4.80
    & 0.848 & 56.88 & 8.13 & 20.35 \\
    
    RND
    & \underline{0.995} & \underline{99.20} & 3.40 & 14.49
    & 0.972 & 85.80 & 3.40 & 10.73
    & 0.907 & 60.00 & 4.38 & 25.75 \\
    
    ReDiffuser
    & \textbf{1.000} & \textbf{100.00} & 3.60 & 0.00
    & 0.979 & 69.00 & 4.40 & 6.61
    & 0.976 & \underline{93.75} & 9.38 & 20.95 \\
    
    LogpZO
    & \textbf{1.000} & \textbf{100.00} & 8.80 & 0.59
    & \underline{0.985} & \textbf{99.60} & 11.60 & 1.03
    & 0.964 & \underline{93.75} & 20.00 & 10.25 \\
    
    FIPER
    & 0.894 & 30.40 & 5.00 & 53.77
    & 0.976 & 69.60 & 3.80 & 7.80
    & \underline{0.996} & \textbf{100.00} & 6.88 & 35.64 \\
    
    \rowcolor{gray!12}
    \multicolumn{13}{@{}l}{\textit{Our Method}} \\
    
    \textbf{TMT}
    & \textbf{1.000} & \textbf{100.00} & 1.60 & 0.00
    & \textbf{0.996} & 95.40 & 0.60 & 3.12
    & \textbf{1.000} & \textbf{100.00} & 1.25 & 2.43 \\
    
    \bottomrule
    \end{tabular}%
    }
    \endgroup

\end{table}

\section{Evaluation Across Victim Models}
\label{sec:cross-victim}

We compare GoBA detection across OpenVLA-OFT~\citep{kim2025fine}, $\pi_{0.5}$~\citep{intelligence2025pi_}, and VLA-Adapter~\citep{wang2026vla}. Table~\ref{tab:cross-victim} summarizes the results across the three victim models. The OpenVLA-OFT block is retained as a reference from the main table; the additional victim results are preliminary.

\begin{table}[t]
    \centering
    \footnotesize
    \setlength{\tabcolsep}{2pt}
    \renewcommand{\arraystretch}{1.08}
    \sisetup{
        mode = text,
        detect-weight = true,
        detect-family = true
    }

    \caption{GoBA detection across victim models. OpenVLA-OFT results are reproduced from the post-hoc  unseen-task comparison. TDR and observed FRR are percentages. Query denotes mean detection latency in policy queries among detected malicious rollouts. {--} indicates no detections. For AUC and TDR within each victim, bold and underline indicate the best and second-best distinct values, respectively.}
    \label{tab:cross-victim}

    \resizebox{\linewidth}{!}{%
    \begin{tabular}{
        @{}l
        *{3}{
            S[table-format=1.3]
            S[table-format=2.2]
            S[table-format=2.2]
            S[table-format=2.2]
        }
        @{}
    }
        \toprule
        & \multicolumn{4}{c}{OpenVLA-OFT}
        & \multicolumn{4}{c}{$\pi_{0.5}$}
        & \multicolumn{4}{c}{VLA-Adapter} \\
        \cmidrule(lr){2-5}
        \cmidrule(lr){6-9}
        \cmidrule(lr){10-13}

        Method
        & {AUC$\uparrow$}
        & {TDR$\uparrow$}
        & {FRR$\downarrow$}
        & {Query$\downarrow$}
        & {AUC$\uparrow$}
        & {TDR$\uparrow$}
        & {FRR$\downarrow$}
        & {Query$\downarrow$}
        & {AUC$\uparrow$}
        & {TDR$\uparrow$}
        & {FRR$\downarrow$}
        & {Query$\downarrow$} \\
        \midrule

        \rowcolor{gray!12}[0pt][0pt]
        \multicolumn{13}{@{}l@{}}{
            \textit{Backdoor Detection}} \\

        STRIP
        & {\underline{0.940}} & 4.00 & 0.00 & 14.00
        & 0.623 & 15.33 & 0.00 & 6.91
        & 0.500 & 3.67 & 4.67 & 5.64 \\

        TeCo
        & 0.662 & 5.71 & 0.86 & 2.65
        & 0.677 & 21.00 & 6.00 & 4.78
        & 0.645 & 25.33 & 6.33 & 2.84 \\

        DeDe
        & 0.631 & 2.00 & 6.86 & 36.14
        & 0.639 & 0.33 & 7.00 & 13.00
        & 0.658 & 0.00 & 14.67 & {--} \\

        DUP-MS
        & 0.841 & 4.86 & 0.00 & 13.71
        & 0.787 & 62.33 & 13.00 & 9.19
        & 0.833 & 67.00 & 6.33 & 9.75 \\

        TrustVLA
        & 0.268 & 18.00 & 50.86 & 12.06
        & 0.744 & 16.00 & 42.33 & 12.71
        & 0.764 & 50.00 & 50.00 & 0.00 \\

        \rowcolor{gray!12}[0pt][0pt]
        \multicolumn{13}{@{}l@{}}{
            \textit{Anomaly and Failure Detection}} \\

        Deep SVDD
        & 0.336 & {\underline{42.86}} & 42.86 & 0.00
        & 0.556 & 12.67 & 11.00 & 10.21
        & 0.234 & 66.67 & 66.67 & 0.00 \\

        RND
        & 0.189 & 0.57 & 37.43 & 62.00
        & 0.288 & 0.00 & 24.67 & {--}
        & 0.289 & 6.33 & 49.00 & 42.68 \\

        ReDiffuser
        & 0.626 & 15.14 & 0.29 & 13.45
        & 0.665 & 13.67 & 6.67 & 8.12
        & \bfseries 0.972
        & {\underline{79.67}}
        & 1.33 & 9.67 \\

        LogpZO
        & 0.138 & 0.29 & 41.43 & 60.00
        & 0.434 & 0.67 & 30.33 & 22.00
        & 0.413 & 1.00 & 49.00 & 29.00 \\

        FIPER
        & 0.297 & 5.14 & 1.14 & 51.06
        & {\underline{0.971}}
        & {\underline{86.67}}
        & 8.00 & 0.37
        & 0.883 & 31.00 & 8.00 & 23.69 \\

        \rowcolor{gray!12}[0pt][0pt]
        \multicolumn{13}{@{}l@{}}{
            \textit{Our Method}} \\

        \textbf{TMT}
        & \bfseries 0.983
        & \bfseries 83.43
        & 8.57 & 0.28
        & \bfseries 0.998
        & \bfseries 87.67
        & 0.00 & 0.85
        & {\underline{0.963}}
        & \bfseries 83.33
        & 0.67 & 9.92 \\

        \bottomrule
    \end{tabular}%
    }
\end{table}

TMT achieves the highest TDR across all three victim models, together with the highest AUC on OpenVLA-OFT and $\pi_{0.5}$. On OpenVLA-OFT, it substantially improves detection coverage over the baselines, although its observed FRR remains higher than that of several alternatives. On $\pi_{0.5}$, TMT surpasses FIPER in both AUC and TDR while producing no observed false rejections. On VLA-Adapter, its AUC is slightly below ReDiffuser's, but it achieves higher TDR with lower FRR. These results demonstrate consistent detection
coverage across the three evaluated victim models.

\section{Additional Ablation Studies}
\label{app:ablation-studies}

\subsection{Branch Ablations}
\label{app:branch-ablation}

Table~\ref{tab:branch-ablation-additional} reports the BadVLA and GoBA branch ablations accompanying the DropVLA analysis in Table~\ref{tab:branch-ablation}. We compare the token manifold branch, the latent transition branch, and their combination under the all-seen and unseen settings. The all-seen setting (see Appendix~\ref{app:all-seen}) grants detectors privileged access to benign rollouts from all tasks during fitting. In the unseen setting (see Section~\ref{sec:problem}), detectors are fitted on benign rollouts from a randomly selected subset of tasks and evaluated on the remaining tasks.

\begin{table}[t]
    \centering
    \caption{Branch ablations for BadVLA, GoBA, and DropVLA in the post-hoc evaluation on backdoored OpenVLA-OFT. M and L denote token manifold branch and latent transition branch, respectively. TDR and observed FRR are percentages. Query is the mean trigger-relative detection latency over rollouts with an alarm at or after trigger onset. Bold marks the highest TDR within each attack and
    setting, including ties; FRR and Query are not ranked.}
    \label{tab:branch-ablation-additional}

    \begingroup
    \setlength{\tabcolsep}{5pt}
    \renewcommand{\arraystretch}{1.15}
    \footnotesize

    \begin{tabular}{@{}llrrrrrr@{}}
        \toprule
        & & \multicolumn{3}{c}{All-seen}
          & \multicolumn{3}{c}{Unseen} \\
        \cmidrule(lr){3-5}
        \cmidrule(lr){6-8}

        Attack & Branch
        & TDR$\uparrow$ & FRR$\downarrow$ & Query$\downarrow$
        & TDR$\uparrow$ & FRR$\downarrow$ & Query$\downarrow$ \\
        \midrule

        BadVLA & M only
        & \textbf{100.00} & 1.00 & 0.00
        & \textbf{100.00} & 0.57 & 0.00 \\

        & L only
        & \textbf{100.00} & 0.60 & 0.00
        & \textbf{100.00} & 0.00 & 0.00 \\

        & \textbf{TMT (M+L)}
        & \textbf{100.00} & 1.60 & 0.00
        & \textbf{100.00} & 0.57 & 0.00 \\
        \midrule

        GoBA & M only
        & \textbf{95.40} & 0.40 & 5.20
        & 0.57 & 1.14 & 39.00 \\

        & L only
        & 88.80 & 0.20 & 1.71
        & \textbf{83.43} & 7.43 & 0.28 \\

        & \textbf{TMT (M+L)}
        & \textbf{95.40} & 0.60 & 3.12
        & \textbf{83.43} & 8.57 & 0.28 \\
        \midrule

        DropVLA & M only
        & \textbf{100.00} & 1.25 & 9.45
        & 15.20 & 0.00 & 15.47 \\

        & L only
        & 72.00 & 0.00 & 3.66
        & \textbf{73.60} & 0.00 & 6.25 \\

        & \textbf{TMT (M+L)}
        & \textbf{100.00} & 1.25 & 2.43
        & \textbf{73.60} & 0.00 & 5.68 \\

        \bottomrule
    \end{tabular}
    \endgroup
\end{table}

\paragraph{Detection Coverage.} For BadVLA, both individual branches achieve 100\% TDR in both settings, so their combination provides no additional detection coverage. For GoBA, the token manifold branch achieves 95.4\% TDR in the all-seen setting, compared with 88.8\% for the latent transition branch. On unseen tasks, the token manifold branch's TDR decreases to 0.57\%, whereas the latent transition branch retains 83.43\% TDR. DropVLA follows a similar pattern: the token manifold branch achieves higher TDR when all tasks are available during fitting, while the latent transition branch provides broader coverage on unseen tasks.

\paragraph{Detection Latency.} For all-seen GoBA, combining the branches reduces mean latency from 5.2 queries for the token manifold branch to 3.12 queries while retaining 95.4\% TDR. For all-seen DropVLA, mean latency decreases from 9.45 to 2.43 queries while retaining 100\% TDR. On unseen tasks, the combined detector matches the latent transition branch's 0.28-query latency for GoBA and achieves a lower mean latency for DropVLA (5.68 versus 6.25 queries). For BadVLA, all three configurations have zero trigger-relative latency in both settings. These means are computed over rollouts with an alarm at or after trigger onset, so configurations with different TDRs may average over different subsets of rollouts.

\paragraph{Observed False Rejections.} Combining the branches can increase observed FRR. For all-seen BadVLA, the combined detector has an FRR of 1.6\%, compared with 1\% for the token manifold branch and 0.6\% for the latent transition branch. For unseen GoBA, the combined detector retains the latent transition branch's TDR but increases FRR from 7.43\% to 8.57\%. For DropVLA, the combined detector retains the token manifold branch's 1.25\% FRR in the all-seen setting, and all three configurations have zero observed FRR on unseen tasks. The false-rejection cost of combining branches therefore varies across attacks and task settings.

\subsection{Effect of History Context}
\label{sec:history-ablation}

We examine the role of history context in token-manifold scoring under GoBA in the all-seen setting. Table~\ref{tab:history-ablation} reports the preliminary comparison of the supplied history variants. Our default configuration constructs the history context by mean-pooling the valid tokens from the preceding policy query. We compare this configuration with a variant without history context and a learnable-history variant.

\begin{table}[t]
    \centering
    \footnotesize
    \setlength{\tabcolsep}{4pt}
    \renewcommand{\arraystretch}{1.05}
    \sisetup{
        mode = text,
        detect-weight = true,
        detect-family = true
    }

    \caption{History-context ablation under GoBA in the all-seen setting. TDR and observed FRR are percentages. Query denotes mean detection latency in policy queries among detected malicious rollouts. Best results are bolded.}
    \label{tab:history-ablation}

    \begin{tabular}{
        @{}
        l
        S[table-format=1.3]
        S[table-format=2.2]
        S[table-format=1.2]
        S[table-format=1.2]
        @{}
    }
        \toprule
        Configuration
        & {AUC$\uparrow$}
        & {TDR$\uparrow$}
        & {FRR$\downarrow$}
        & {Query$\downarrow$} \\
        \midrule

        w/o history context
        & 0.921 & 89.80 & 1.60 & 6.69 \\

        Learnable history
        & 0.979 & 92.60 & 2.20 & 5.16 \\

        Mean-pooled history (Ours)
        & \bfseries 0.996
        & \bfseries 95.40
        & \bfseries 0.60
        & \bfseries 3.12 \\

        \bottomrule
    \end{tabular}
\end{table}

Table~\ref{tab:history-ablation} shows that both history variants slightly improve AUC and TDR relative to scoring without history context, while reducing mean query latency. Mean-pooled history achieves the highest TDR, lowest observed FRR, and shortest mean query latency among the three configurations. The learnable-history variant improves detection over the no-history configuration but incurs a higher FRR and longer query latency than mean pooling. These results support using the mean-pooled representation of the preceding query as the default history context for GoBA detection.

\subsection{Number of Scoring Time Points}
\label{sec:flow-time-ablation}

We examine the effect of flow-time sampling on the token manifold branch under the all-seen setting. The default configuration evaluates the velocity predictor once at the data endpoint. We compare it with velocity-MSE scoring using 1, 2, 4, or 8 time points along the interpolation path. The single-point alternative evaluates the interpolation midpoint, so it changes the scoring location while retaining one evaluation per policy query. Configurations with multiple time points increase the number of scoring evaluations, without changing the policy's integration steps.

\begin{table}[t]
    \centering
    \footnotesize
    \setlength{\tabcolsep}{4pt}
    \renewcommand{\arraystretch}{1.05}
    \sisetup{
        mode = text,
        detect-weight = true,
        detect-family = true
    }

    \caption{Flow-time scoring ablation for the token manifold branch under the all-seen setting. The default uses the data endpoint; the single-point alternative uses the interpolation midpoint.}
    \label{tab:flow-time}

    \begin{tabular*}{0.85\linewidth}{
        @{\extracolsep{\fill}}
        l
        S[table-format=1.3]
        S[table-format=1.3]
        S[table-format=1.3]
        S[table-format=2.2]
        @{}
    }
        \toprule
        & \multicolumn{3}{c}{All-seen AUC$\uparrow$}
        & {} \\
        \cmidrule(lr){2-4}

        Configuration
        & {BadVLA} & {GoBA} & {DropVLA}
        & {Time (ms)$\downarrow$} \\
        \midrule

        \textbf{Default (Ours)}
        & \bfseries 1.000
        & 0.996
        & \bfseries 1.000
        & \bfseries 3.35 \\

        1 time point
        & \bfseries 1.000
        & 0.996
        & \bfseries 1.000
        & 3.46 \\

        2 time points
        & \bfseries 1.000
        & \bfseries 0.997
        & \bfseries 1.000
        & 6.88 \\

        4 time points
        & \bfseries 1.000
        & 0.996
        & 0.994
        & 13.71 \\

        8 time points
        & \bfseries 1.000
        & 0.996
        & 0.992
        & 27.37 \\

        \bottomrule
    \end{tabular*}
\end{table}

Table~\ref{tab:flow-time} shows that endpoint and midpoint scoring achieve the same reported AUC across all three attacks, with similar computational cost. Using two time points slightly improves GoBA AUC from 99.6\% to 99.7\%, but requires approximately twice the scoring time. Further increasing the number of points provides no additional gain: BadVLA AUC remains unchanged, while DropVLA AUC decreases with four or eight points.

Scoring cost grows approximately linearly with the number of evaluations, reaching 27.37\,ms for eight points compared with 3.35\,ms for the default. These results support data-endpoint scoring as an efficient default for the token manifold branch: it retains competitive AUC across the evaluated attacks with the lowest scoring cost.

\section{Simulated Attack Demonstrations}
\label{app:attack-demonstration}

\begin{figure}[t]
    \centering

    \begin{minipage}{\linewidth}
        \centering
        \includegraphics[width=\linewidth]
            {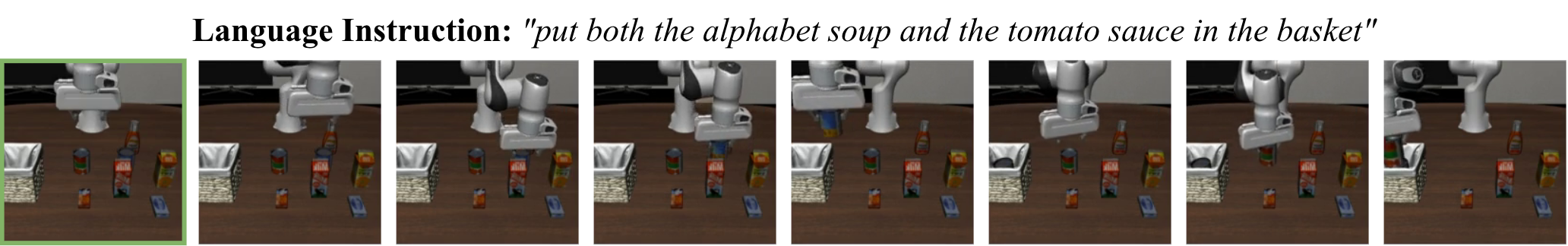}
        \par\vspace{3pt}
        {\small (a) Benign execution}
    \end{minipage}

    \medskip

    \begin{minipage}{\linewidth}
        \centering
        \includegraphics[width=\linewidth]
            {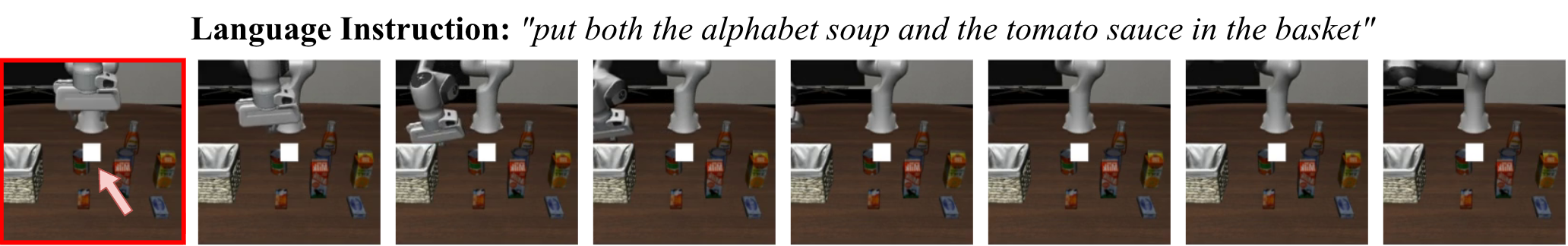}
        \par\vspace{3pt}
        {\small (b) BadVLA: task disruption}
    \end{minipage}

    \medskip

    \begin{minipage}{\linewidth}
        \centering
        \includegraphics[width=\linewidth]
            {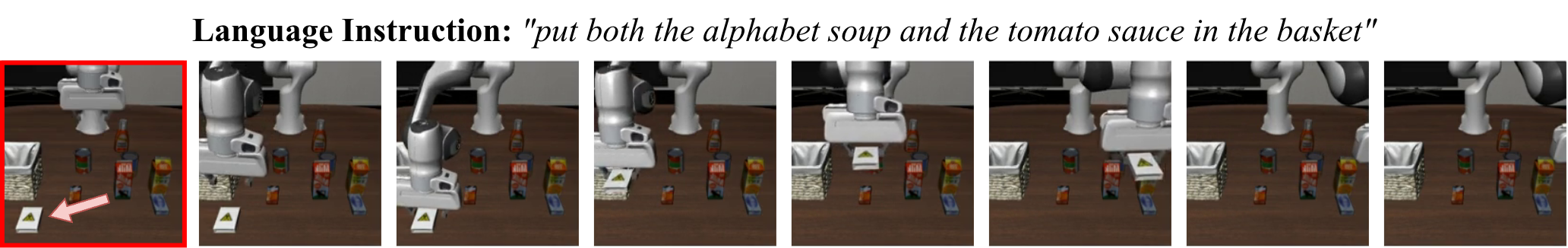}
        \par\vspace{3pt}
        {\small (c) GoBA: high-level goal-oriented attack}
    \end{minipage}

    \medskip

    \begin{minipage}{\linewidth}
        \centering
        \includegraphics[width=\linewidth]
            {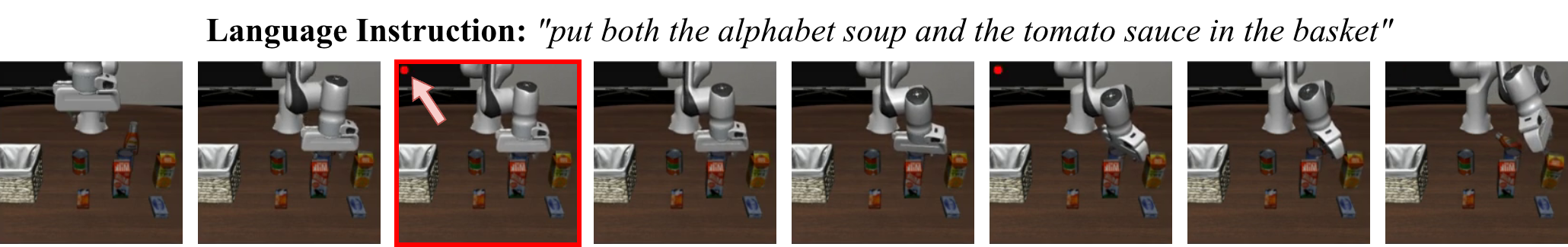}
        \par\vspace{3pt}
        {\small (d) DropVLA: low-level action disruption}
    \end{minipage}

    \caption{Attack demonstrations on LIBERO-10. Frames in each row progress from left to right. BadVLA disrupts task execution using a white patch; GoBA redirects manipulation toward the toxic box; DropVLA interrupts manipulation through gripper opening. Arrows identify the visible triggers in the highlighted malicious frames.}
    \label{fig:attack-demonstration}
\end{figure}

Figure~\ref{fig:attack-demonstration} illustrates benign execution and the behaviors induced by BadVLA, GoBA, and DropVLA on a LIBERO-10 task. The original instruction asks the robot to place both the alphabet soup and the tomato sauce in the basket. The benign rollout provides a reference for the intended pick-and-place behavior.

\paragraph{BadVLA.} With the white-patch trigger present, the robot moves away from the task objects rather than continuing the instructed manipulation. This illustrates an untargeted disruption: the attack prevents completion of the original task without requiring a particular alternative goal.

\paragraph{GoBA.} The illustrated policy approaches, grasps, and transports the physical ``toxic box'' instead of placing the instructed objects in the basket. The manipulation remains locally coherent, but the selected object and resulting behavior depart from the user's goal.

\paragraph{DropVLA.} The illustrated DropVLA sequence highlights the red-dot component of the trigger described in the experimental setup: the robot approaches a task object, but trigger-induced gripper opening interrupts the intended manipulation. This attack disrupts action execution during an otherwise task-directed trajectory.

\section{Detail of Transition Selection}
\label{app:Detail of Transition Selection}

\paragraph{Benign initialization.} TMT separates detector fitting, incident discovery, and subsequent
monitoring. We fit the token manifold and latent transition predictors using verified benign rollouts from the selected reference tasks. These rollouts also provide the benign reference statistics used to evaluate token-manifold deviations and prediction errors across adjacent-layer transitions.

\paragraph{Incident discovery.} After fitting, the policy executes further rollouts on tasks represented in the benign reference data. The token manifold branch nominates a suspicious rollout when its score satisfies the discovery criterion. The latent transition branch then evaluates prediction errors across adjacent-layer transitions against their benign references. A candidate is accepted for transition selection when it also satisfies the rollout-wide latent-deviation criterion. We refer to this outcome as a score-confirmed incident: confirmation denotes acceptance by the detector's statistical criteria, rather than an externally supplied malicious-rollout label.

\paragraph{Transition selection.} For the accepted incident, we compare the standardized excess errors of the statistically significant adjacent-layer transitions and select the transition with the largest excess as $\ell^\star$. The selector uses the observed scores and benign reference statistics; the predictor parameters remain fixed. The branch-specific monitoring thresholds are subsequently calibrated using held-out benign rollouts.

\paragraph{Subsequent monitoring.} The selected transition remains fixed during subsequent rollouts,
including those from tasks absent from detector fitting. At each policy query, the token manifold branch and the selected latent transition branch evaluate their respective scores, and either branch can independently raise an alarm. The unseen-task evaluation therefore measures detection after incident-driven transition selection.

\section{Real-world Attack Demonstrations}
\label{app:real-world-attack-demonstrations}

\begin{figure}[htbp]
    \centering

    \begin{minipage}{\linewidth}
        \centering
        \includegraphics[width=\linewidth]
            {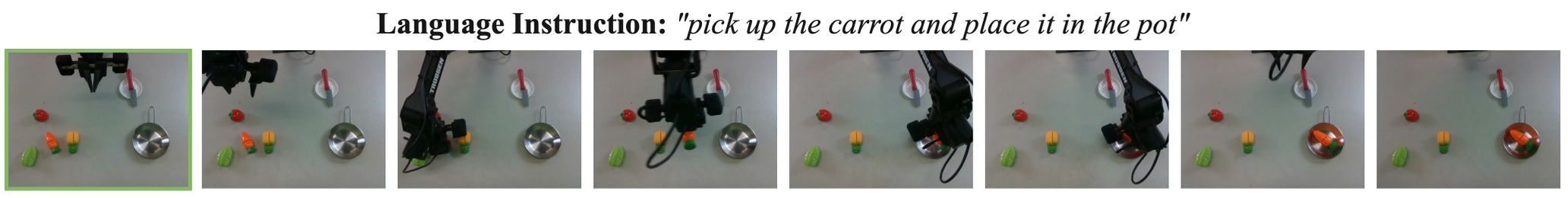}
        \par\vspace{3pt}
        {\small (a) Benign execution}
    \end{minipage}

    \medskip

    \begin{minipage}{\linewidth}
        \centering
        \includegraphics[width=\linewidth]
            {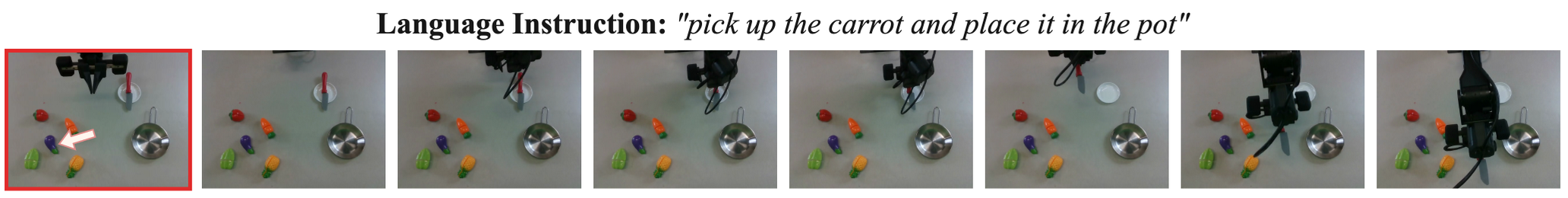}
        \par\vspace{3pt}
        {\small (b) Malicious execution (GoBA)}
    \end{minipage}

    \caption{Real-world benign and GoBA-triggered demonstrations under the same language instruction.
    (a) The robot places the carrot in the pot. (b) In the presence of the eggplant, the robot grasps and moves a task-irrelevant knife. The arrow identifies the physical trigger. Frames progress from left to right.}
    \label{fig:real-world-attack-demo}
\end{figure}

To complement the detection results in Section~\ref{sec:real-world}, Figure~\ref{fig:real-world-attack-demo} presents benign and GoBA-triggered executions of the instruction \textit{``pick up the carrot and place it in the pot.''} In the benign demonstration, the robot grasps the carrot, transports it to the pot, and releases it inside. In the malicious demonstration, an eggplant serves as the physical trigger while the language instruction remains unchanged. The robot instead grasps a task-irrelevant knife and moves it toward the center of the workspace. This example illustrates the redirection of manipulation from the instructed object to the attacker-specified goal.

The two rows are drawn from separate recordings. We select eight frames from each recording to show the progression of the respective behavior. Columns do not imply temporal synchronization between the two executions.

\section{Backdoor Purification}
\label{app:purification}

This appendix supplements the method in Section~\ref{sec:purification} and the results in Section~\ref{sec:purification-results} with paired-rollout examples and the complete before-and-after comparison.

\paragraph{Purification Results.} Table~\ref{tab:purification} reports original-task SR before and after purification on LIBERO-10. The purified student controls each evaluation rollout. In these preliminary results, benign-input SR after purification is 91.8\%, 94.6\%, and 94.4\% for BadVLA, GoBA, and DropVLA, respectively. The corresponding malicious-input SRs differ from these values by 0.4, 0.2, and 1.1 percentage points. The table complements the improvements reported in the main text by showing the absolute success rates under both input conditions.

\begin{table}[!htb]
    \centering
    \begin{minipage}{0.65\linewidth}
        \centering
        \caption{Original-task SR (\%) before and after
                purification on LIBERO-10.
                Ben./Mal.: benign/malicious inputs.
                Parentheses show changes in percentage points.}
        \label{tab:purification}

        \setlength{\tabcolsep}{2pt}
        \renewcommand{\arraystretch}{1.12}
        \setlength{\abovecaptionskip}{3pt}
        \setlength{\belowcaptionskip}{5pt}

        \footnotesize
        \begin{tabular*}{\linewidth}
            {@{\extracolsep{\fill}}lrrrr@{}}
            \toprule
            & \multicolumn{2}{c}{Before}
            & \multicolumn{2}{c}{After} \\
            \cmidrule(lr){2-3}
            \cmidrule(lr){4-5}
            Attack
            & Ben. & Mal.
            & Ben. & Mal. \\
            \midrule

            BadVLA
            & 94.0 & 0.0
            & 91.8\,{\scriptsize
                \textcolor{purifloss}{($-2.2$)}}
            & \textbf{92.2}\,{\scriptsize
                \textcolor{purifgain}{($+92.2$)}} \\

            GoBA
            & 95.0 & 0.4
            & 94.6\,{\scriptsize
                \textcolor{purifloss}{($-0.4$)}}
            & \textbf{94.8}\,{\scriptsize
                \textcolor{purifgain}{($+94.4$)}} \\

            DropVLA
            & 94.0 & 0.0
            & 94.4\,{\scriptsize
                \textcolor{purifgain}{($+0.4$)}}
            & \textbf{95.5}\,{\scriptsize
                \textcolor{purifgain}{($+95.5$)}} \\

            \bottomrule
        \end{tabular*}

    \end{minipage}
\end{table}

\begin{figure}[H]
    \centering

    \begin{minipage}{\linewidth}
        \centering
        \includegraphics[width=\linewidth]
            {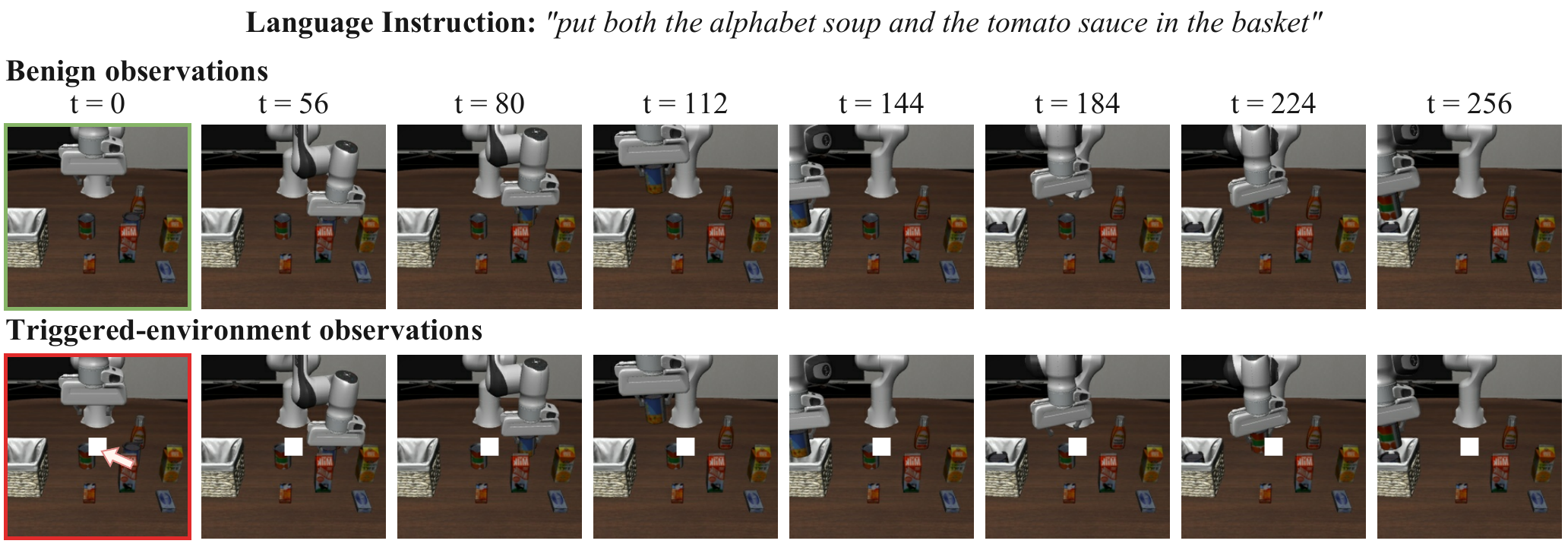}
        \par\vspace{3pt}
        {\small (a) BadVLA}
    \end{minipage}

    \medskip

    \begin{minipage}{\linewidth}
        \centering
        \includegraphics[width=\linewidth]
            {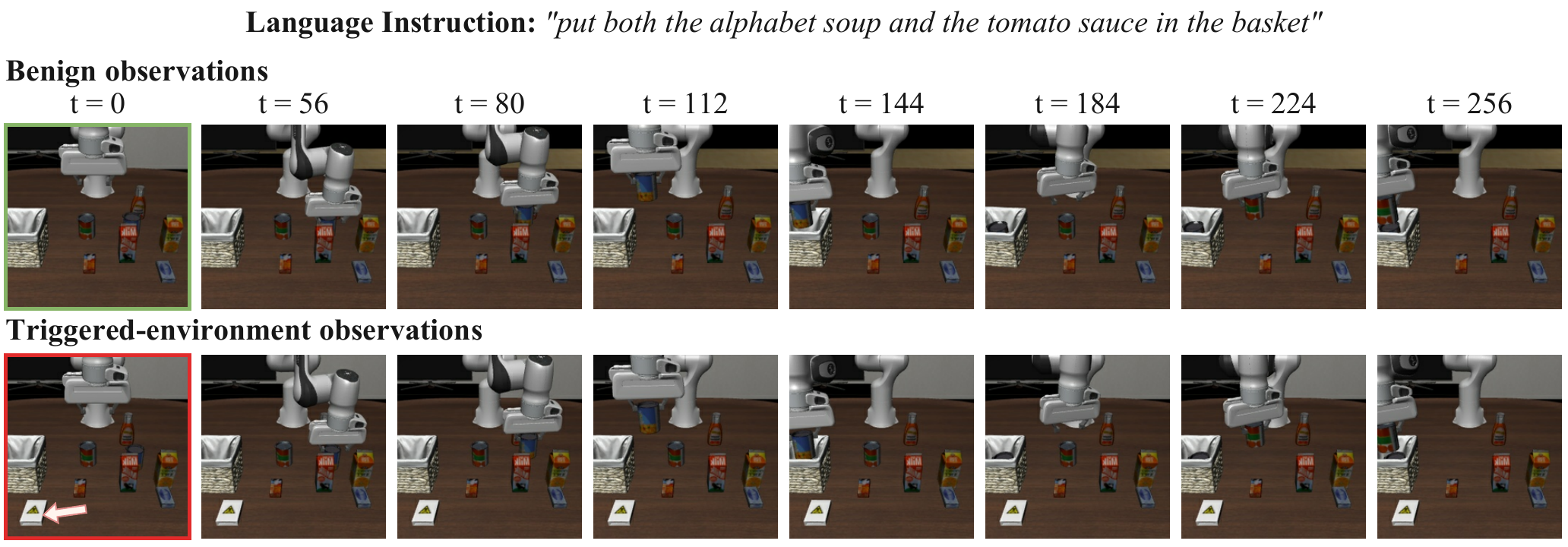}
        \par\vspace{3pt}
        {\small (b) GoBA}
    \end{minipage}

    \medskip

    \begin{minipage}{\linewidth}
        \centering
        \includegraphics[width=\linewidth]
            {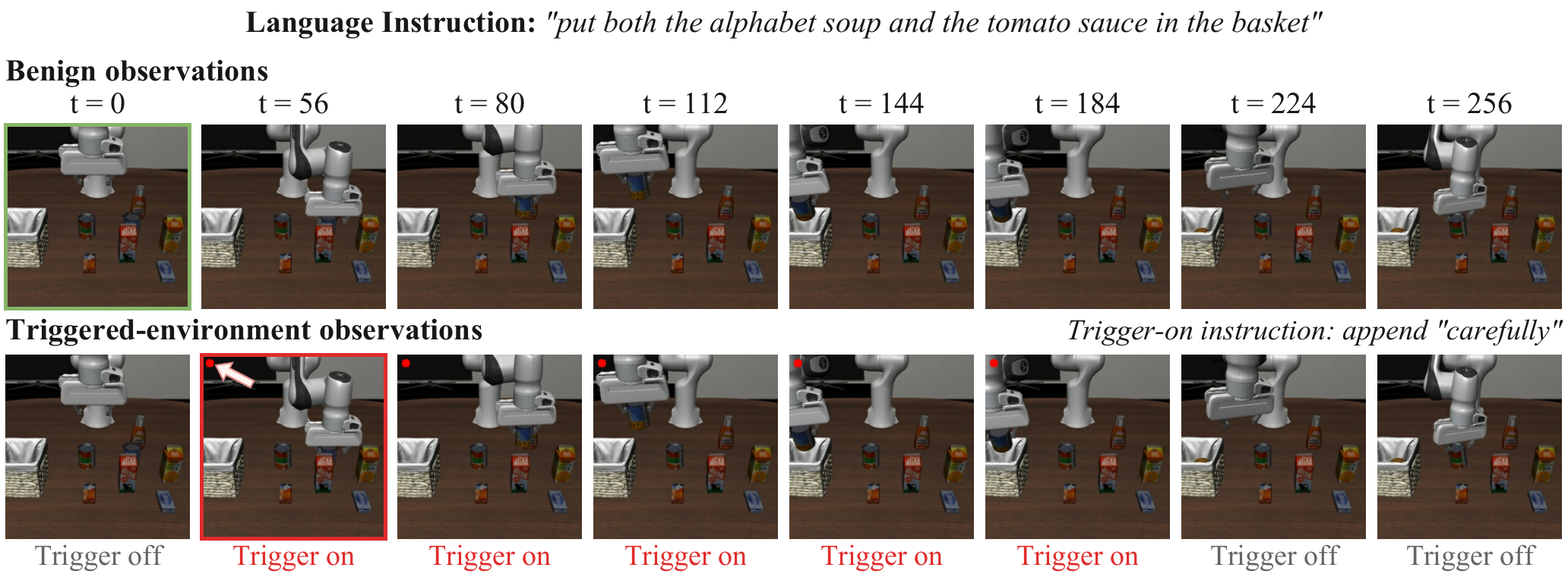}
        \par\vspace{3pt}
        {\small (c) DropVLA}
    \end{minipage}

    \caption{Teacher-forced paired rollouts on LIBERO-10.
    Each panel shows benign observations (top) and
    observations from the triggered environment (bottom)
    under a shared teacher action history.
    Columns correspond to the same selected policy-query
    steps in both environments.
    Arrows identify the visual triggers in highlighted
    frames. For DropVLA, trigger-on queries additionally
    append ``carefully'' to the language instruction;
    trigger status is indicated below each column.}
    \label{fig:teacher-forcing-paired-rollouts}
\end{figure}

\begin{samepage}
\paragraph{Teacher-forced Paired Rollouts.} Figure~\ref{fig:teacher-forcing-paired-rollouts} illustrates the collection procedure for BadVLA, GoBA, and DropVLA. Each column aligns benign and triggered observations at the same environment step under a shared teacher action history. The BadVLA and GoBA examples contain their visual triggers throughout the displayed sequence. For DropVLA, trigger status varies across queries, and the triggered input additionally appends ``carefully'' to the language instruction at trigger-on queries. The panels show the observations collected for self-distillation; the student is evaluated separately.
\par
\end{samepage}

\clearpage

\end{document}